\documentclass[10pt,twocolumn,letterpaper]{article}

\usepackage[pagenumbers]{iccv} 

\definecolor{iccvblue}{rgb}{0.21,0.49,0.74}
\usepackage[pagebackref,breaklinks,colorlinks,allcolors=iccvblue]{hyperref}

\usepackage{adjustbox}
\usepackage{bm}
\usepackage{multirow}
\usepackage{multicol}
\usepackage{color, colortbl}
\usepackage{pifont}
\usepackage{placeins}
\newcommand{\au}[1]{\setlength{\fboxsep}{1pt}\colorbox{gold}{#1}}
\newcommand{\ag}[1]{\setlength{\fboxsep}{1pt}\colorbox{silver}{#1}}
\newcommand{\cu}[1]{\setlength{\fboxsep}{1pt}\colorbox{bronze}{#1}}
\usepackage[capitalize]{cleveref}
\Crefname{section}{Section}{Sections}
\crefname{section}{Sec.}{Secs.}
\Crefname{align}{Equation}{Equations}
\crefname{align}{Eq.}{Eqs.}
\Crefname{equation}{Equation}{Equations}
\crefname{equation}{Eq.}{Eqs.}
\Crefname{figure}{Figure}{Figures}
\crefname{figure}{Fig.}{Figs.}
\Crefname{table}{Table}{Tables}
\crefname{table}{Tab.}{Tabs.}

\newcommand\minisection[1]{\vspace{1mm}\noindent \textbf{#1}}
\newcommand{\norm}[1]{\left\lVert#1\right\rVert}

\newcommand{\cmark}{\ding{51}}
\newcommand{\xmark}{\ding{55}}

\newcommand{\fullmodel}{\mbox{Total-Editing}\xspace}
\newcommand{\supp}{\mbox{Supp.~Mat.}\xspace}
\newcommand{\supst}{$^\text{st}$\xspace}
\newcommand{\supnd}{$^\text{nd}$\xspace}
\newcommand{\suprd}{$^\text{rd}$\xspace}
\newcommand{\supth}{$^\text{th}$\xspace}
\definecolor{Gray}{gray}{0.9}
\definecolor{Celadon}{rgb}{0.67, 0.88, 0.69}
\definecolor{Cream}{rgb}{1.0, 0.99, 0.82}

\definecolor{gold}{HTML}{FAE37F}
\definecolor{silver}{HTML}{D7D7D7}
\definecolor{bronze}{HTML}{EDBA91}

\DeclareMathOperator{\argmin}{argmin}

\newcommand{\Dc}{\mathcal{D}}
\newcommand{\Ec}{\mathcal{E}}
\newcommand{\Fc}{\mathcal{F}}

\newcommand{\Lc}{\mathcal{L}}

\newcommand{\Rc}{\mathcal{R}}

\newcommand{\Tc}{\mathcal{T}}

\newcommand{\Rb}{\mathbb{R}}

\newcommand{\av}{\mathbf{a}}

\newcommand{\cv}{\mathbf{c}}
\newcommand{\dv}{\mathbf{d}}
\newcommand{\ev}{\mathbf{e}}

\newcommand{\lv}{\mathbf{l}}

\newcommand{\nv}{\mathbf{n}}
\newcommand{\ov}{\mathbf{o}}
\newcommand{\pv}{\mathbf{p}}

\newcommand{\rv}{\mathbf{r}}
\newcommand{\sv}{\mathbf{s}}
\newcommand{\tv}{\mathbf{t}}

\newcommand{\vv}{\mathbf{v}}
\newcommand{\wv}{\mathbf{w}}

\newcommand{\zv}{\mathbf{z}}

\newcommand{\Cv}{\mathbf{C}}

\newcommand{\Fv}{\mathbf{F}}

\newcommand{\Iv}{\mathbf{I}}

\newcommand{\Lv}{\mathbf{L}}

\newcommand{\Nv}{\mathbf{N}}

\newcommand{\Sv}{\mathbf{S}}
\newcommand{\Tv}{\mathbf{T}}

\newcommand{\Vv}{\mathbf{V}}

\newcommand{\Xv}{\mathbf{X}}

\def\paperID{1766} 
\def\confName{ICCV}
\def\confYear{2025}

\title{Total-Editing: Head Avatar with Editable Appearance, Motion, and Lighting}

\author{
\centering
\begin{tabular}{c}
Yizhou Zhao\textsuperscript{1} \quad
Chunjiang Liu\textsuperscript{1} \quad
Haoyu Chen\textsuperscript{1} \quad
Bhiksha Raj\textsuperscript{1} \quad
Min Xu\textsuperscript{1} \\
Tadas Baltrušaitis\textsuperscript{2} \quad
Mitch Rundle\textsuperscript{2} \quad
HsiangTao Wu\textsuperscript{2} \quad
Kamran Ghasedi\textsuperscript{2} \\
\textsuperscript{1}Carnegie Mellon University \quad
\textsuperscript{2}Microsoft
\end{tabular}
}

\begin{document}

\twocolumn[{%
\renewcommand\twocolumn[1][]{#1}%
\maketitle
\begin{center}
    \centering
    \vspace{-1.5em}
    \includegraphics[width=\linewidth]{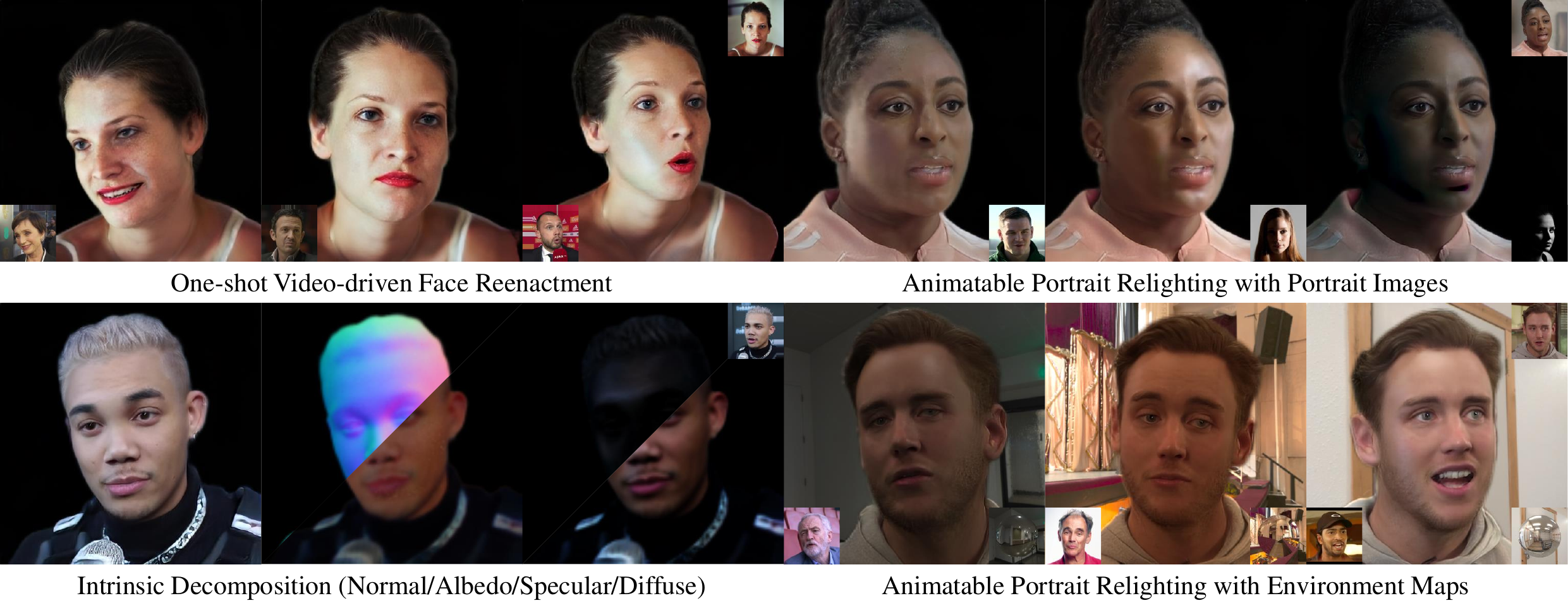}
    \captionsetup{type=figure}
    \vspace{-1.5em}
    \caption{Our \fullmodel enables geometry-and-illumination-aware portrait editing with appearance sources (top-right corner), motion sources (bottom-left corner), and lighting sources (bottom-right corner) through intrinsically decomposed neural radiance fields.}
    \label{fig:teaser}
\end{center}
}]

\begin{abstract}
Face reenactment and portrait relighting are essential tasks in portrait editing, yet they are typically addressed independently, without much synergy. Most face reenactment methods prioritize motion control and multiview consistency, while portrait relighting focuses on adjusting shading effects. To take advantage of both geometric consistency and illumination awareness, we introduce \fullmodel, a unified portrait editing framework that enables precise control over appearance, motion, and lighting. Specifically, we design a neural radiance field decoder with intrinsic decomposition capabilities. This allows seamless integration of lighting information from portrait images or HDR environment maps into synthesized portraits. We also incorporate a moving least squares based deformation field to enhance the spatiotemporal coherence of avatar motion and shading effects. With these innovations, our unified framework significantly improves the quality and realism of portrait editing results. Further, the multi-source nature of \fullmodel supports more flexible applications, such as illumination transfer from one portrait to another, or portrait animation with customized backgrounds.
\end{abstract}
\vspace{-1.0em}    
\section{Introduction}
\label{sec:intro}

Generating realistic human portraits is essential for various applications, including virtual reality, augmented reality, social media, gaming, and film production.
Within these fields, face reenactment and portrait relighting are two pivotal tasks.
Face reenactment involves transferring motion, i.e., facial expressions and head pose, from one person to another, and enables lifelike talking-head applications such as video conferencing~\cite{wang2020one,google2021project}, virtual avatars~\cite{google2021project,thies2019neural}, and video dubbing~\cite{yu2023dinet,agarwal2022audio}.
Portrait relighting, on the other hand, focuses on modifying a portrait’s lighting to match diverse environments, enhancing realism and immersion by adapting to the dynamic lighting in virtual spaces~\cite{wenger2005performance,sun2021nelf,yeh2022learning}.
Despite their inherent synergy, face reenactment and portrait relighting have largely been treated as separate tasks.

This raises a question, \textit{should they be considered jointly?}
Indeed, face reenactment benefits from dynamic relighting to ensure natural light and shadow transitions during head motion, while portrait relighting can leverage the abundant monocular video datasets used in reenactment, which are more accessible than traditional light-stage datasets.
Jointly addressing both tasks improves adaptability and realism in portrait editing, opening new possibilities for immersive applications.
To this end, we introduce \fullmodel, a novel portrait editing framework that integrates face reenactment and portrait relighting in an end-to-end pipeline.

\begin{figure}[t]
    \begin{center}
    \includegraphics[width=\linewidth]{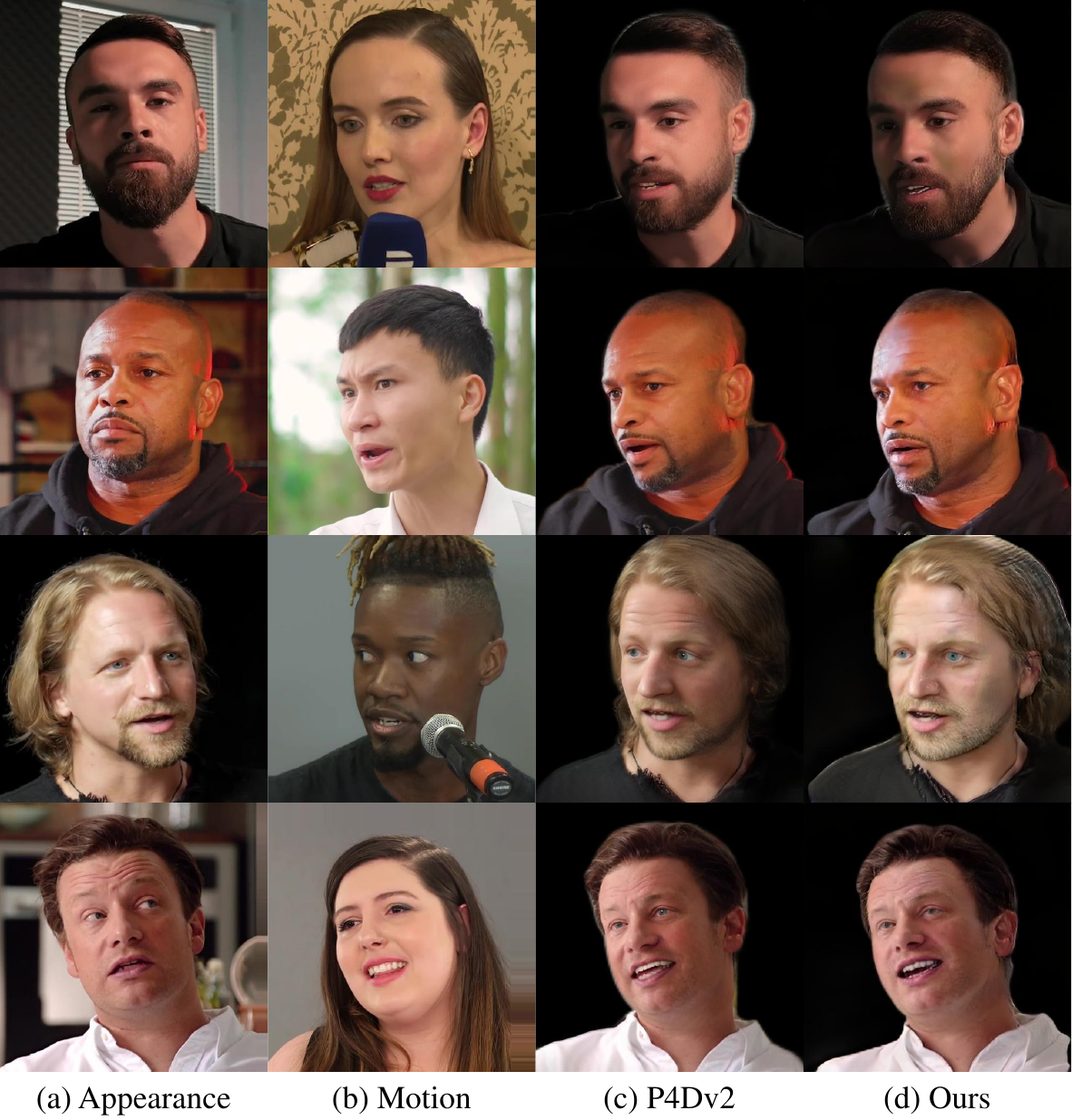}
    \vspace{-1.5em}
    \caption{\textbf{Face reenactment under uneven lighting.} Existing models like~\cite{deng2024portrait4dv2} couple facial textures and lighting, resulting in fixed light and shadow that do not adapt to head movements. In contrast, our model provides more realistic portrait shading.} 
    \label{fig:qualitative_uneven_lighting}
    \end{center}
    \vspace{-2em}
\end{figure}

Our design is guided by several key insights.  
First, existing face reenactment methods~\cite{deng2024portrait4d,deng2024portrait4dv2,ye2024real3d,li2024generalizable,chu2024gpavatar} struggle to model lighting variations without explicit guidance, leading to fixed light and shadow under uneven lighting conditions, as shown in~\cref{fig:qualitative_uneven_lighting}. 
This challenge becomes even more pronounced when the training data lacks large head motions or strong shading effects, which are essential for learning illumination consistency.
To address this, we introduce an intrinsically decomposed neural radiance field 
(NeRF) decoder to decompose volumetric color into intrinsic components, allowing direct control over lighting.
Moreover, we develop a physically rendered dataset that captures subject motion under diverse lighting conditions to complement real data and enhance illumination awareness.

Next, 3D Morphable Models (3DMM)~\cite{blanz1999morphable} based motion editing~\cite{deng2024portrait4d, deng2024portrait4dv2} relies on surface fields (SF)~\cite{bergman2022generative} for feature warping.
Due to the reliance on nearest neighbors, it often requires additional spatio-temporal smoothing to mitigate discontinuities in feature propagation.
To this end, we propose a Moving Least Squares (MLS)~\cite{schaefer2006image,zhu20073d} based deformation field, which naturally ensures smooth and continuous deformations through its globally aware MLS kernel.
Unlike SF-based methods, MLS supports both translational deformation, which adjusts only position attributes, and rotational deformation, which is crucial for transforming directional properties like normal vectors.
This supports our model to provide more realistic portrait shading, as in~\cref{fig:qualitative_uneven_lighting}.

Finally, \fullmodel achieves portrait synthesis with disentangled appearance, motion, and lighting conditions. Shown in~\cref{fig:teaser}, our method produces natural light shifts on the face during head motion (top-left), robust results with lighting estimated from portraits (top-right); realistic HDR environment map based illumination (bottom-right); and intrinsic decomposition into specular, diffuse, normal and albedo components (bottom-left).

Our contributions can be summed up as follows:
\begin{itemize}
    \item We present a novel framework, \fullmodel, that conducts 3D-aware portrait generation given an appearance source, a motion source, and a lighting source, facilitating more precise control and more versatile applications compared to face reenactment and portrait relighting models.
    \item We introduce an intrinsically decomposed NeRF decoder that uses estimated or pre-filtered lightmaps to represent lighting conditions. This allows flexible lighting control of portraits via source images or HDR environment maps.
    \item We propose an MLS-based deformation field, which supports general affine deformation, including translation and rotation. It produces improved spatiotemporal consistencies than its SF-based counterparts. 
    \item We contribute a physically rendered synthetic dataset for general portrait editing, featuring 2M frames that capture diverse subjects, views, poses, expressions, and lighting environments. Each subject is presented with corresponding motion sequences across multiple environments.
\end{itemize}

\section{Related Work}
\label{sec:related}
\minisection{Face Reenactment} has seen significant advancements in recent years initially built on the success of CNN and GAN~\cite{lecun1998gradient,goodfellow2014generative,karras2020analyzing}. Early strategies inserted features from driving images into 2D generative networks to create animatable portraits~\cite{zakharov2019few,mildenhall2020nerf,wang2023progressive}. Recent approaches represent expressions and head poses as warp fields, deforming source images to match driving images~\cite{siarohin2019first,ren2021adaptive,hong2022depth}. While these methods produce high-quality images, they often lack 3D consistency, limiting realistic results under varied poses and expressions. Some methods incorporate 3D Morphable Models (3DMM) \cite{blanz1999morphable} into 2D frameworks~\cite{khakhulin2022realistic,ma2023otavatar,li2023one,li2024generalizable,yu2023nofa,chu2024gpavatar,chu2024generalizable}, but they are still limited by the accuracy of monocular 3DMM reconstructions.
Building on the success of neural radiance fields (NeRF)~\cite{mildenhall2020nerf}, several methods~\cite{gafni2021dynamic,park2021nerfies,tretschk2021nonrigid,athar2022rignerf,park2021hypernerf} have adopted NeRF for head reconstruction, but their reliance on multi-view or single-view videos limits generalization. Some works \cite{xu2023pv3d,tang20233dfaceshop,zhuang2022controllable,xu2023omniavatar} train generators for controllable head avatars based on identity inversion, but often fail to preserve the source identity due to inversion limitations. 
By learning canonical triplane representations in NeRF-based models, Trevithick et al.~\cite{trevithick2023real} provide real-time 3D-aware novel view synthesis without expression change via volume rendering, and Portrait4D series~\cite{deng2024portrait4d,deng2024portrait4dv2} tackles dynamic expression modeling using synthetic and pseudo multi-view data. Despite their advancements, these methods have difficulty separating lighting effects from facial features, leading to fixed light and shadows on the face. In contrast, our approach achieves precise illumination consistency, allowing natural lighting shifts in response to head movements.

\begin{figure*}[t]
    \begin{center}
    \includegraphics[width=\linewidth]{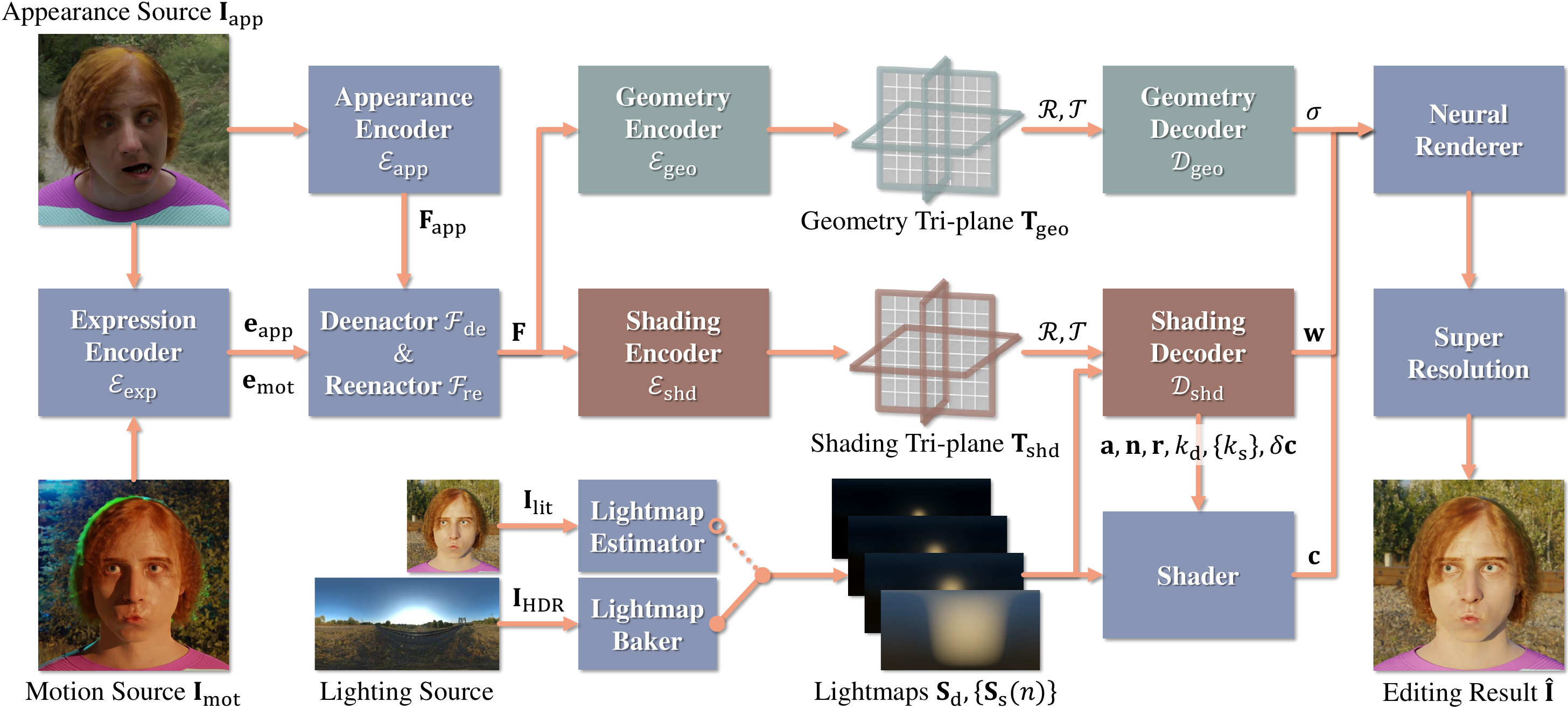}
    \caption{\textbf{The framework of \fullmodel.} \cref{subsec:appearance_motion}: Our pipeline learns to encode appearance and motion sources $\Iv_\text{app},\Iv_\text{mot}$, neutralize the expression from $\Iv_\text{app}$ and reapply the expression from $\Iv_\text{mot}$ to obtain a fused feature $\Fv$. After generating canonical space geometry and shading tri-planes $\Tv_\text{geo},\Tv_\text{shd}$, the neck pose is handled by warping features with moving least squares based deformation fields $\Rc,\Tc$. \cref{subsec:illumination}: For the lighting source, we either estimate from a portrait image $\Iv_\text{lit}$ or pre-filter (bake) an HDR environment map $\Iv_\text{HDR}$, resulting in diffuse and specular lightmaps $\Sv_\text{d},\{\Sv_\text{s}(n)\}$. \cref{subsec:geometry_shading}: With the lighting information, geometry and shading decoders $\Dc_\text{geo},\Dc_\text{shd}$ decode point-wise attributes. Finally, a neural renderer and a super-resolution module render the editing result $\hat{\Iv}$.}
    \label{fig:pipeline}
    \end{center}
    \vspace{-2em}
\end{figure*}

\minisection{Portrait Relighting} aims to realistically re-illuminate human face images. Early work by Debevec et al.~\cite{debevec2000acquiring} introduced a method for HDR face relighting using a light stage to capture images one light at a time (OLAT). This method was extended by~\cite{wenger2005performance,sun2019single,wang2020single,zhang2021neural}, but they are limited to subjects captured within the light stage setup. To address this, \cite{yeh2022learning,fei2023split,papantoniou2023relightify,zhou2019deep} utilize synthetic data for training. Some advancements in portrait relighting leverage diverse approaches, including diffusion models, neural fields, GANs, and physics-guided models~\cite{caselles2023sira,sun2021nelf,hou2022face,hou2021towards,kim2024switchlight,nestmeyer2020learning,ren2024relightful,mei2024holo,rao2024lite2relight,abdal2021styleflow}. For lighting representations, some approaches~\cite{saito2024relightable,qiu2024relitalk,jiang2023nerffacelighting,deng2024lumigan,ranjan2023facelit,zhou2019deep,sun2019single,ponglertnapakorn2023difareli,mei2024holo,guo2025high} utilize Spherical Harmonics (SH)~\cite{ramamoorthi2001efficient} as a compact lighting representation. Other methods~\cite{lin2024edgerelight360, yeh2022learning, pandey2021total, ji2022geometry, mei2023lightpainter,rao2024lite2relight} use pre-filtered lightmaps from HDR environment maps to capture higher frequency lighting details. Recently, SwitchLight~\cite{kim2024switchlight} leverages the Cook-Torrance reflectance model~\cite{cook1982reflectance} for precise light-surface interactions. Unlike these approaches which are mostly 2D-based, our method leverages 3D representations for free-view rendering and video-conditioned animations, requiring less training data and facilitating more flexible applications.

\section{Method}
\label{sec:method}
Taking as input an appearance source $\Iv_\text{app}$, a motion source $\Iv_\text{mot}$, and a lighting source, either a portrait image $\Iv_\text{lit}$ or an HDR environment map $\Iv_\text{HDR}$, our goal is to synthesize a 3D head that combines the appearance of $\Iv_\text{app}$, the motion of $\Iv_\text{mot}$, and the lighting of $\Iv_\text{lit}$ or $\Iv_\text{HDR}$. To achieve this, we present the \fullmodel framework, as depicted in~\cref{fig:pipeline}, disentangling the control of appearance, motion, and lighting. For appearance, we use a pre-trained appearance encoder~\cite{deng2024portrait4dv2} to extract appearance features from $\Iv_\text{app}$. For motion, we employ an off-the-shelf expression encoder~\cite{wang2023progressive} to extract appearance-free expression features, allowing us to neutralize the expression in $\Iv_\text{app}$ (de-enactment) and apply the expression from $\Iv_\text{mot}$ (re-enactment), and moving least squares based deformation fields to capture the neck pose. As for lighting, we leverage pre-filtered HDR environment maps, i.e., lightmaps, as our lighting representation. Further, we decompose point-wise colors in 3D volumes with the Phong reflection model to isolate shading effects from portrait materials. In this way, our framework effectively disentangles and integrates appearance, motion, and lighting conditions, enabling the synthesis of realistic 3D head models with precise control over each attribute.

\subsection{Preliminaries}
\subsubsection{Phong Reflection Model}
The Phong reflection model~\cite{phong1998illumination} is a widely used lighting model for simulating the way surfaces reflect light. For each point on the surface, it decomposes the reflection into three terms, namely, the ambient reflection ($\cv_\text{a}$), the diffuse reflection ($\cv_\text{d}$), and the specular reflection ($\cv_\text{s}$),
\begin{align}
    \cv = \cv_\text{a} + \cv_\text{d} + \cv_\text{s},
\end{align}
where $\cv$ is the total reflection intensity at this point. We omit the ambient component $\cv_\text{a}$ in our model and calculate the diffuse and specular components as
\begin{align}
\label{eq:int_diffuse}
    \cv_\text{d} &= k_\text{d}\av\odot\sv_\text{d},
    \;\sv_\text{d}=\int_\Omega{\Lv(\lv)\left(\nv\cdot\lv\right) d\lv}, \\
\label{eq:int_specular}
    \cv_\text{s} &= \sum_n k_\text{s}(n)\sv_\text{s}(n),
    \;\sv_\text{s}(n)=\int_\Omega{\Lv(\lv)(\rv\cdot\lv)^n d\lv}.
\end{align}
Here, $\odot$ denotes the Hadamard product, scalars $k_\text{d}$ and $k_\text{s}(n)$ are the diffuse and specular coefficients, $\av$ is the surface albedo, and $n\in\{1,16,32,64\}$ is the shininess exponent. We refer to $\sv_\text{d}$ and $\sv_\text{s}(n)$ as diffuse and specular shadings. They sum the incoming light $\Lv(\lv)$ from all directions $\lv$ over the hemisphere $\Omega$ above the surface. In the integral, $\nv$ and $\rv$ are the surface normal and the reflected viewing direction, with
\begin{align}
\label{eq:reflect}
    \rv = 2(\nv\cdot\vv)\nv - \vv,
\end{align}
where $\vv$ is the viewing direction.

\subsubsection{Pre-filtering Environment Maps}
\label{subsubsec:prefilter}
Inspired by~\cite{heidrich1999realistic,kautz2000unified}, we pre-filter HDR environment maps with Phong lobes to avoid expensive real-time reflection computations. For each surface normal $\nv$ in~\cref{eq:int_diffuse}, we pre-integrate the diffuse shading as $\sv_\text{d}(\nv)$. Similarly, for each reflected viewing direction $\rv$ in~\cref{eq:int_specular}, we precompute the specular shading as $\sv(n,\rv)$. Aggregating each gives us the diffuse lightmap $\Sv_\text{d}=\{\sv_\text{d}(\nv)\}_{\nv\in\Omega}$ and specular lightmaps $\{\Sv_\text{s}(n)\}_n=\{\{\sv_\text{s}(n,\rv)\}_{\rv\in\Omega}\}_n$. This process is also referred to as lightmap baking. At runtime, shading calculations are simplified to look-ups in these precomputed lightmaps with $\nv$ or $\rv$, reducing the computational complexity of rendering each pixel from $O(N)$, where $N$ is the number of incident rays from the environment, to $O(1)$.

\subsubsection{Image Formation}
Integrating the Phong reflection model into volumetric rendering, we derive the expected color $\Cv(\pv)$ of camera ray $\pv(t)=\ov+t\dv$ with $t\in[t_n,t_f]$ and $\dv=-\vv$,
\begin{align}
    \Cv(\pv)=\int_{t_n}^{t_f}T(t)\sigma(\pv(t))\cv(\pv(t),\vv)dt,
    \\
    T(t)=\exp\left(-\int_{t_n}^{t}\sigma(\pv(s))ds\right),
\end{align}
where $T$ is the accumulated transmittance and $\sigma$ is the volume density. We rewrite reflection as $\cv(\pv(t),\vv)$ since its diffuse component $\cv_\text{d}$ is position dependent and its specular component $\cv_\text{s}$ is position and view dependent,
\begin{align}
    \cv_\text{d} &= \cv_\text{d}(\pv) = k_\text{d}(\pv)\av(\pv)\odot\sv_\text{d}(\nv(\pv)),
    \\
    \cv_\text{s} &= \cv_\text{s}(\pv,\vv) = \sum_n k_\text{s}(n,\pv)\sv_\text{s}(n,\nv(\pv),\vv).
\end{align}

\subsection{Learning Appearance and Motion}
\label{subsec:appearance_motion}

\begin{figure}[t]
    \begin{center}
    \includegraphics[width=\linewidth]{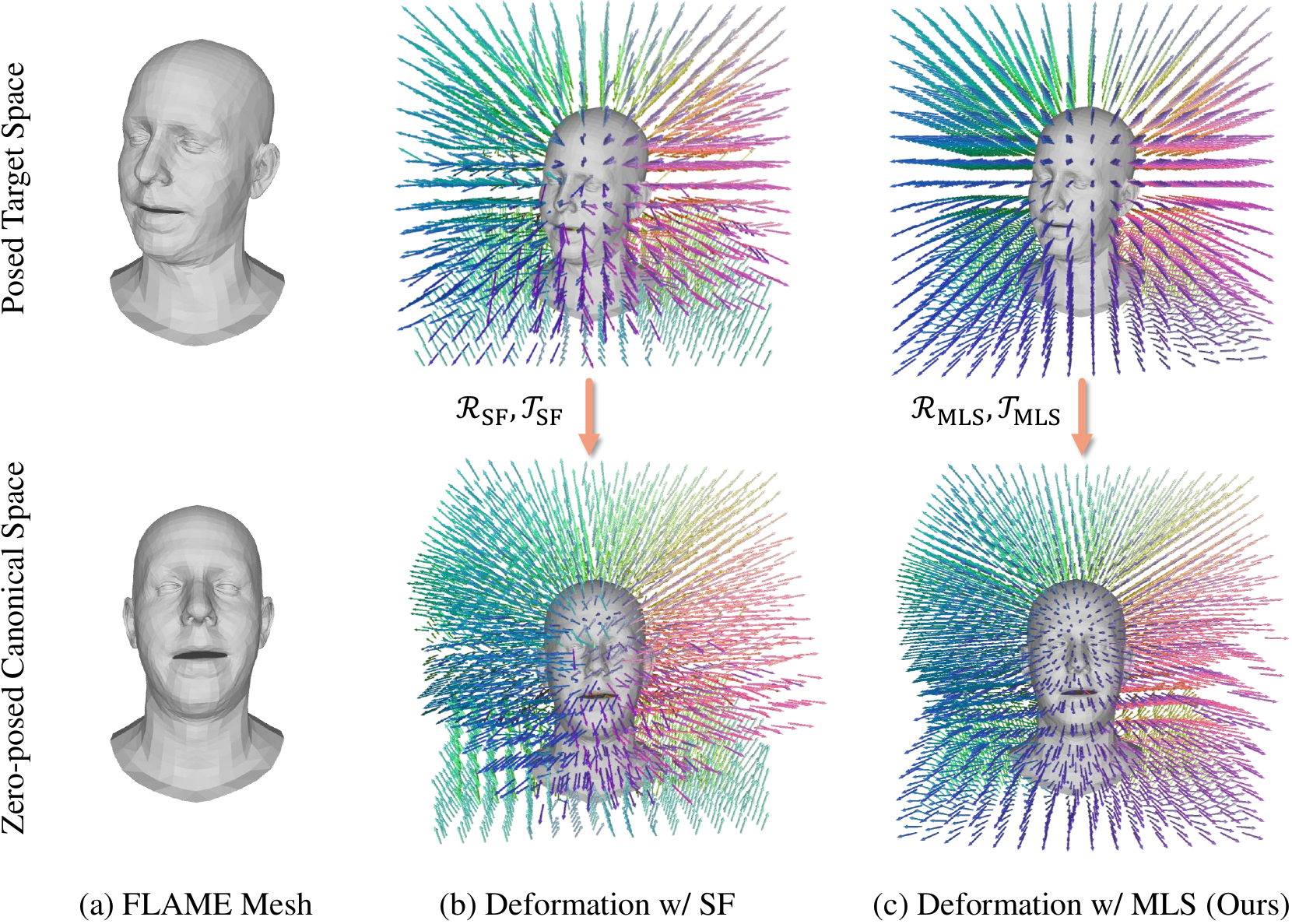}
    \caption{\textbf{Deformation field comparison.} (a) Similar to~\cite{deng2024portrait4d,deng2024portrait4dv2} we derive deformation field from FLAME meshes. Points and attached normals are sampled from the posed target space and then deformed into the unposed canonical space by deformation fields $\Tc$ and $\Rc$. (b) Surface Field (SF) based approach~\cite{bergman2022generative} assigns each grid to the nearest mesh triangle, leading to discontinuous deformation results. (c) In contrast, our moving least squares (MLS) based deformation field weighs per-point deformation with its inverse distance to all mesh vertices, producing smoother results.} 
    \label{fig:deformation}
    \end{center}
    \vspace{-1.5em}
\end{figure}

We follow~\cite{deng2024portrait4d,deng2024portrait4dv2} to incorporate appearance and motion control in \fullmodel with decoupled learning. Specifically, we adopt an off-the-shelf expression encoder $\Ec_\text{mot}$ to extract 1D expression features for both appearance and motion sources, $\ev_\text{app} = \Ec_\text{mot}(\Iv_\text{app})$, $\ev_\text{mot} = \Ec_\text{mot}(\Iv_\text{mot})$,
and an appearance encoder $\Ec_\text{app}$ to extract 2D appearance features from the appearance source, $\Fv_\text{app} = \Ec_\text{app}(\Iv_\text{app})$.
All these features are fused with two Transformer-based modules, a deenactor $\Fc_\text{de}$ and a following reenactor $\Fc_\text{re}$,
\begin{align}
    \Fv &= \Fc_\text{re}(\Fc_\text{de}(\Fv_\text{app}, \ev_\text{app}), \ev_\text{mot}).
\end{align}
Together, they learn to embed the appearance from $\Iv_\text{app}$, neutralize it, and inject the expressions from $\Iv_\text{mot}$. Using the fused appearance and expression features $\Fv$, we generate a geometry tri-plane $\Tv_\text{geo}$ and a shading tri-plane $\Tv_\text{shd}$,
\begin{align}
    \Tv_\text{geo} &= \Ec_\text{geo}(\Fv),\;\Tv_\text{shd} = \Ec_\text{shd}(\Fv),
\end{align}
where $\Ec_\text{geo}$ and $\Ec_\text{shd}$ are the geometry and shading encoders, respectively, both with a ViT-based architecture~\cite{trevithick2023real}. To handle the neck pose, we further derive deformation fields $\Tc,\Rc$ with FLAME meshes. They will then be utilized to warp tri-plane features and rotate decoded normals from the unposed canonical space to the posed target space, as shown in~\cref{fig:deformation}. Unlike~\cite{deng2024portrait4d,deng2024portrait4dv2} using the Surface Field (SF) approach~\cite{bergman2022generative} that determines the deformation for each sample point by the motion of nearest triangles, we adopt moving least squares (MLS)~\cite{zhu20073d} instead to obtain continuous deformation results. Our intuition is that by globally averaging transformations, MLS enables more natural transitions and reduces the risk of artifacts that can occur with the more localized SF method. Specifically, we set posed mesh vertices $\Vv^\text{t}$ and their unposed correspondences $\Vv^\text{c}$ as control points, and solve for the transformation $\Tc$ of sample point $\pv$ by
\begin{align}
    \argmin_\Tc \sum_i w_i(\Vv^\text{t}, \pv)\norm{\Tc(\Vv^\text{t}_i)-\Vv^\text{c}_i}_2^2,
\end{align}
where the weights are of the form
\begin{align}
    w_i(\Xv, \pv)=\norm{\Xv_i-\pv}_2^{-2\alpha},
\end{align}
with $\alpha=1.0$ being a fall-off parameter. Note that this formulation lets $w_i(\Vv^\text{t}, \Vv^\text{t}_i)=\infty$ and thus $\Tc(\Vv^\text{t}_i)=\Vv^\text{c}_i$. Confining $\Tc$ to be a rigid transformation, this minimization can be solved via singular value decomposition. Similarly, we obtain the rotation field $\Rc$ by controlling with normals
\begin{align}
    \argmin_\Rc \sum_i w_i(\Vv^\text{t}, \pv)\norm{\Rc(\Nv^\text{t}_i)-\Nv^\text{c}_i}_2^2,
\end{align}
where $\Nv^\text{t}$ and $\Nv^\text{c}$ are surface normal directions on posed and unposed meshes, respectively.

\subsection{Learning Geometry and Shading}
\label{subsec:geometry_shading}

\begin{figure}[t]
    \begin{center}
    \includegraphics[width=\linewidth]{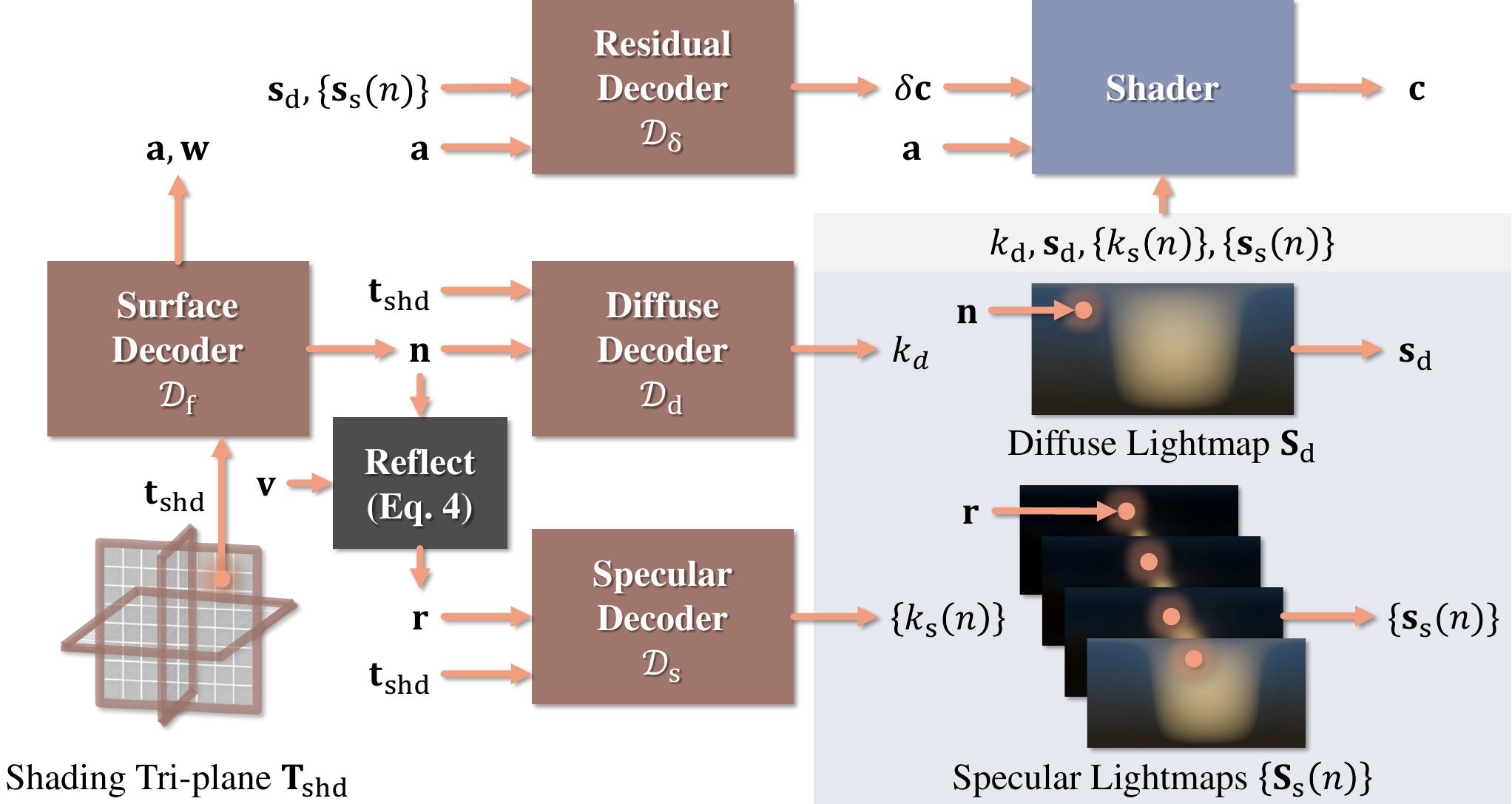}
    \caption{\textbf{Shading module architecture.} The shading feature $\tv_\text{shd}$ sampled from a position in the shading tri-plane is first decoded into normal $\nv$, albedo $\av$, and additional features $\wv$ for super-resolution. The viewing direction $\vv$ is then reflected with the normal $\nv$, resulting in a reflected viewing direction $\rv$. $\nv$ and $\rv$ are concatenated with $\tv_\text{shd}$ and mapped to a diffuse coefficient $k_\text{d}$ and specular coefficients $\{k_\text{s}(n)\}$ for various shininess values $\{n\}$, respectively. They are also used to sample a diffuse shading $\sv_\text{d}$ and specular shadings $\{\sv_\text{s}(n)\}$ from corresponding lightmaps. These shadings, $\sv_\text{d},\{\sv_\text{s}(n)\}$, are concatenated with the albedo $\av$ to decode a residual color $\delta\cv$. The final color $\cv$ at this position is obtained by combining PBR and neural residual $\delta\cv$ in~\cref{eq:render}.} 
    \label{fig:shading}
    \end{center}
    \vspace{-1.5em}
\end{figure}

We then decode point-wise attributes with MLP-based decoders. Specifically, for each sample point $\pv$ in the geometry tri-plane, the geometry decoder $\Dc_\text{geo}$ decode its volume density from the geometry feature $\tv_\text{geo}$ at this point,
\begin{align}
    \sigma &= \Dc_\text{geo}(\tv_\text{geo}),\;\text{where }\tv_\text{geo}=\Tv_\text{geo}(\Tc(\pv)),
\end{align}
and $\Tc$ is the FLAME-derived translation field warping $\pv$ from the target space to the canonical space. To inject illumination awareness into our neural renderer, we extend the color decoder in~\cite{chan2022efficient} to a shading decoder $\Dc_\text{shd}$ that decomposes the volume color according to the Phong reflection model. It incorporates a surface decoder $\Dc_\text{f}$, a diffuse decoder $\Dc_\text{d}$, a specular decoder $\Dc_\text{s}$, and a residual decoder $\Dc_\delta$, as illustrated in~\cref{fig:shading}. From the shading feature $\tv_\text{shd}$ at point $\pv$, the surface decoder $\Dc_\text{f}$ decodes canonical space normal $\nv^\text{c}$, albedo $\av$, and additional features $\wv$ for super-resolution,
\begin{align}
    \nv^\text{c}, \av, \wv &= \Dc_\text{f}(\tv_\text{shd}),\;\text{where }\tv_\text{shd}=\Tv_\text{shd}(\Tc(\pv)).
\end{align}
Different from $\sigma,\av,\wv$ which are the same in zero pose and target pose, the canonical normal needs to be rotated to the target normal $\nv=\Rc^{-1}(\nv^\text{c})$ with the target-to-canonical rotation field $\Rc$. \cref{eq:reflect} reflects the normal $\nv$ with the viewing direction $\vv$ of the sampled ray to obtain a reflected viewing direction $\rv$. Then, the diffuse decoder $\Dc_\text{d}$ and the specular decoder $\Dc_\text{s}$ predict shading coefficients by utilizing these surface attributes with the shading feature $\tv_\text{shd}$,
\begin{align}
    k_\text{d} &= \Dc_\text{d}([\tv_\text{shd},\nv]),\;\{k_\text{s}(n)\} = \Dc_\text{s}([\tv_\text{shd},\rv]),
\end{align}
where $[\cdot,\cdot]$ denotes a concatenation. We also sample diffuse and specular lightmaps $\Sv_\text{d},\{\Sv_\text{s}(n)\}$ with the normal $\nv$ and the reflected viewing direction $\rv$, respectively, resulting in diffuse and specular shadings $\sv_\text{d},\{\sv_\text{s}(n)\}$. To enhance the realism beyond naive Phong shading, we further decode a residual color $\delta\cv$ with a residual decoder $\Dc_\delta$,
\begin{align}
    \delta\cv &= \Dc_\delta([\av,\sv_\text{d},\{\sv_\text{s}(n)\}]).
\end{align}
This allows our shader to combine physically based rendering (PBR) with learnable residuals as the volume color,
\begin{align}
\label{eq:render}
    \cv &= k_\text{d}\av\odot\sv_\text{d} + \sum_n k_\text{s}(n)\sv_\text{s}(n) + \delta\cv.
\end{align}
Finally, we feed point-wise density $\sigma$, color $\cv$, and extra features $\wv$ to a neural renderer and a subsequent super-resolution module to obtain editing result $\hat{\Iv}$.

\subsection{Learning Illumination}
\label{subsec:illumination}

\begin{figure}[t]
    \begin{center}
    \includegraphics[width=\linewidth]{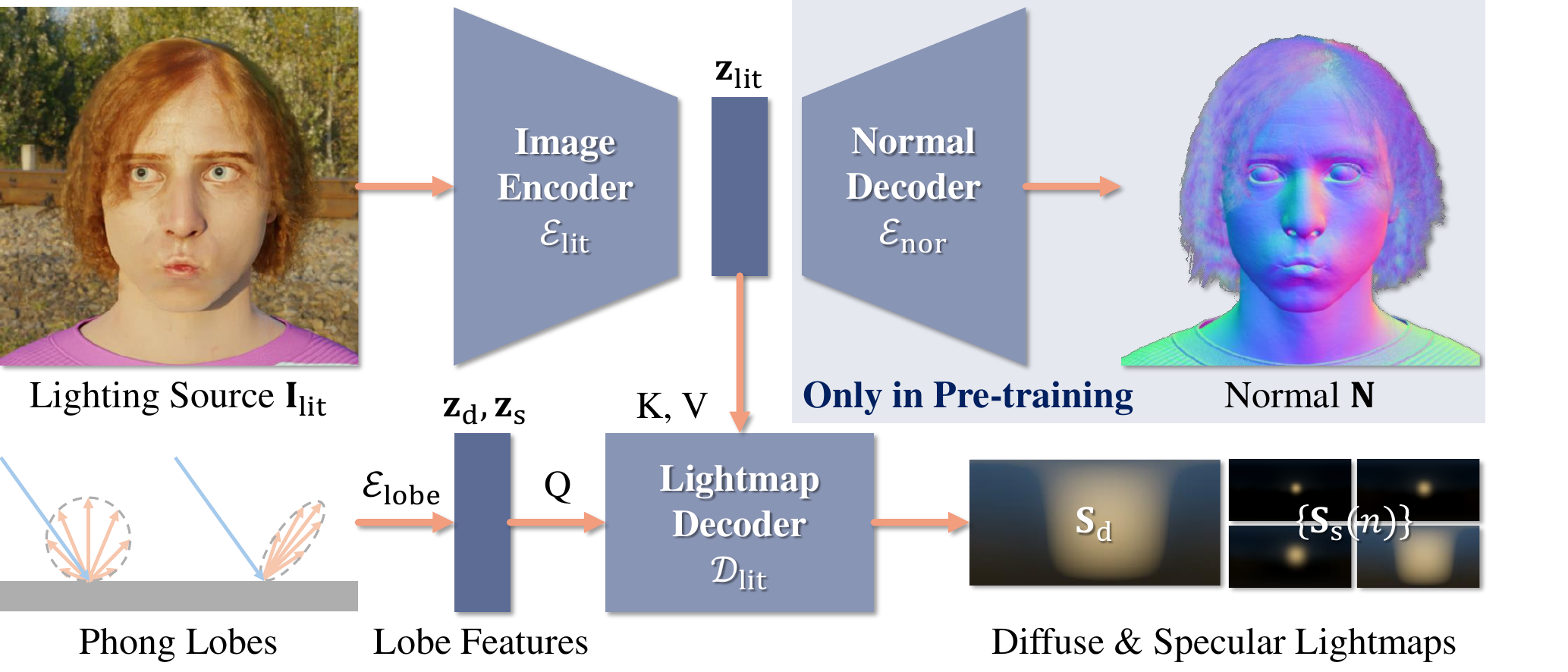}
    \caption{\textbf{Lightmap estimator architecture.} During pre-training with synthetic data, we adopt a U-Net to encode lighting source $\Iv_\text{lit}$ and decode pixel-wise normal $\Nv$. The intermediate feature $\zv_\text{lit}$ is used to decode diffuse and specular lightmaps $\Sv_\text{d},\{\Sv_\text{s}(n)\}$, querying with embedded lobe features $\zv_\text{d},\{\zv_\text{s}(n)\}$. In joint training with \fullmodel, the normal decoder is detached.}
    \label{fig:lightmap}
    \end{center}
    \vspace{-1.5em}
\end{figure}

With this intrinsic decomposition of our 3D portrait, we can illuminate it using either a portrait image, $\Iv_\text{lit}$, or an HDR environment map, $\Iv_\text{HDR}$. Both lighting sources are converted into lightmaps for sampling by our shader. In addition to the conversion of environment maps introduced in~\cref{subsubsec:prefilter}, we further present our lightmap estimator inspired by~\cite{kim2024switchlight} to predict lightmaps from portrait images. As shown in~\cref{fig:lightmap}, it leverages a U-Net architecture for encoding lighting source $\Iv_\text{lit}$ and decoding its pixel-wise normal $\Nv$. To estimate lightmaps $\Sv_\text{d},\{\Sv_\text{s}(n)\}\in\Rb^{H \times W \times 3}$, diffuse and specular Phong lobes are projected into lobe features $\zv_\text{d},\{\zv_\text{s}(n)\}\in\Rb^{H \times W \times C}$ with a shared linear layer $\Ec_\text{lobe}$,
\begin{align}
    \zv_\text{d} &= \Ec_\text{lobe}\left(\max(0, \nv\cdot\lv)\right), \\
    \zv_\text{s}(n) &= \Ec_\text{lobe}\left(\max(0, (\rv\cdot\lv)^n)\right),
\end{align}
where $\nv,\rv\in[-1,1]^{H \times W \times 3}$ are query normals and reflected viewing directions, respectively, and $\lv\in[-1,1]^{N \times 3}$ are quantized light directions over a sphere. Then, a Transformer-based Lightmap decoder $\Dc_\text{lit}$ queries the encoded portrait image $\zv_\text{lit}$ with these lobe features, producing diffuse and specular lightmaps,
\begin{align}
    \Sv_\text{d} = \Dc_\text{lit}(\zv_\text{d},\zv_\text{lit}),\;\Sv_\text{s}(n)=\Dc_\text{lit}(\zv_\text{s}(n),\zv_\text{lit}).
\end{align}

\section{Experiments}
\label{sec:exp}

\subsection{Experimental Setting}
\label{subsec:setting}
\minisection{Datasets.}
We jointly train our \fullmodel with synthetic and real data. For the synthetic dataset, we render a multi-view subset and a video-like one. The former includes 50K subjects rendered in 2 environments, with each subject captured from 10 different views. The latter contains 10K subjects rendered across 10 environments, each with 10 varied poses and expressions. Each unique view, pose, or expression is shared across all environments. The whole dataset with roughly 2M images in total will be released to benefit the community. More details available in the \supp For real data, we use the VFHQ~\cite{xie2022vfhq} dataset, which comprises 15K video clips. We also evaluate our model with its 100-clip test split. All data are at resolution $512^2$.

\minisection{Evaluation Metrics.}
We employ common metrics to evaluate both the synthesis quality and control accuracy of all baseline methods and our model. For image quality, we use PSNR, SSIM, Fréchet Inception Distances (FID) ~\cite{heusel2017gans}, and LPIPS~\cite{zhang2018unreasonable} to assess perceptual similarity and distribution alignment between generated images and ground truths. To measure identity preservation, we calculate the cosine similarity (CSIM) between the face recognition features~\cite{deng2019arcface} of generated images and the appearance sources. For expression and pose control accuracy, we use Average Expression Distance (AED) and Average Pose Distance (APD)~\cite{lin20223d}, derived from a 3DMM estimator~\cite{deng2019accurate}.

\minisection{Implementation Details.}
During pre-training of the lightmap estimator with synthetic data, we utilize the Adam optimizer with a learning rate of $1 \times 10^{-4}$ and a batch size of $32$, optimizing  $\Ec_\text{lit},\Dc_\text{lit},\Ec_\text{nor}$ for 500K steps.
Subsequently, we plug the lightmap estimator into \fullmodel.
Upon the pre-trained  $\Ec_\text{app},\Ec_\text{mot},\Fc_\text{de},\Fc_\text{re},\Ec_\text{geo},\Dc_\text{geo}$ from Portrait4D-v2~\cite{deng2024portrait4dv2}, we learn $\Fc_\text{re},\Ec_\text{geo},\Dc_\text{geo},\Ec_\text{shd},\Dc_\text{shd},\Ec_\text{lit},\Dc_\text{lit}$ using a combination of synthetic data and VFHQ-Train, while keeping the remaining components of \fullmodel fixed.
In this phase, training proceeds for 1M steps with an Adam optimizer, a learning rate of $1 \times 10^{-4}$, and a batch size of $12$.
During training, we randomly sample appearance source $\Iv_\text{app}$ and motion source $\Iv_\text{mot}$ of one subject, with the editing target $\Iv^*$ equal to the motion source $\Iv_\text{mot}$. As for the lighting source, we use the HDR environment map of $\Iv_\text{mot}$ for synthetic data and another random frame from the same video for real data.
More details can be found in \supp

\subsection{Comparison Results}
\begin{table*}[t]
\centering
\footnotesize
\setlength{\tabcolsep}{0.4em}
\adjustbox{width=0.9\linewidth}{
\begin{tabular}{ll|cccccc|cccc}
    \toprule
    \multicolumn{2}{c|}{Method} & \multicolumn{6}{c|}{Self Reenactment} & \multicolumn{4}{c}{Cross Reenactment} \\
    Reenactment & Relighting & PSNR $\uparrow$ & SSIM $\uparrow$ & LPIPS $\downarrow$ & CSIM $\uparrow$ & AED $\downarrow$ & APD $\downarrow$ & CSIM $\uparrow$ & AED $\downarrow$ & APD $\downarrow$ & FID $\downarrow$ \\
    \midrule
        \multirow{2}{*}{GPAvatar~\cite{chu2024gpavatar}} & $-$ & \au{20.7} & \au{0.753} & \cu{0.256} & 0.802 & \cu{0.176} & \ag{0.021} & 0.517 & 0.383 & 0.037 & 55.5 \\
        & Cai et al.~\cite{cai2024real} & \cu{19.5} & 0.699 & 0.330 & 0.524 & 0.239 & 0.028 & 0.372 & \cu{0.381} & 0.044 & 59.8 \\
    \midrule
        \multirow{2}{*}{Real3DPortrait~\cite{ye2024real3d}} & $-$ & 19.3 & \cu{0.711} & 0.270 & \cu{0.856} & 0.217 & \cu{0.026} & \ag{0.685} & 0.434 & 0.044 & \cu{46.2} \\
        & Cai et al.~\cite{cai2024real} & 18.6 & 0.681 & 0.340 & 0.561 & 0.263 & 0.035 & 0.443 & 0.424 & 0.055 & 50.2 \\
    \midrule
        \multirow{2}{*}{Portrait4D-v2~\cite{deng2024portrait4dv2}} & $-$ & 18.9 & 0.704 & \ag{0.247} & \ag{0.874} & \ag{0.154} & 0.027 & \cu{0.686} & 0.386 & \ag{0.031} & \ag{39.6} \\
        & Cai et al.~\cite{cai2024real} & 18.2 & 0.665 & 0.337 & 0.552 & 0.232 & 0.027 & 0.430 & \ag{0.380} & \cu{0.034} & 43.9 \\
    \midrule
        \multicolumn{2}{c|}{\fullmodel (Ours)} & \ag{20.3} & \ag{0.730} & \au{0.226} & \au{0.896} & \au{0.148} & \au{0.015} & \au{0.713} & \au{0.370} & \au{0.024} & \au{38.3} \\
    \bottomrule
\end{tabular}
}
\caption{\textbf{Comparison results on VFHQ-Test at resolution $512^2$.} We use colors to denote \au{first}, \ag{second}, and \cu{third} places, respectively.}
\vspace{-1.0em}
\label{tab:main}
\end{table*}

In~\cref{tab:main}, we first compare \fullmodel with other one-shot video-based face reenactment methods directly. For baseline methods, we test for standard self-reenactment and cross-reenactment settings, where an appearance source is given, and a motion source is the video of the same subject or another subject, respectively. As for our \fullmodel, we also use the appearance source as the lighting source to let the editing result reflect the source lighting situation. From empirical results in~\cref{fig:qualitative_uneven_lighting}, we observe current face reenactment models often couple facial textures with lighting, resulting in unrealistic fixed shading effects on generated results. We thus try to inject illumination into the reenacted faces with a state-of-the-art portrait relighting method~\cite{cai2024real}, forming two-stage reenactment-relighting pipelines. However, this leads to performance degradation. The reason might be 1) accumulated errors that propagate from the reenacted faces to the relighting model, and 2) the SH lighting representation used in~\cite{cai2024real} capturing mainly diffuse shadings. Overall, our end-to-end method outperforms all other approaches in the face reenactment task. 

\subsection{Ablation study}

\begin{table}[t]
\centering
\footnotesize
\setlength{\tabcolsep}{0.4em}
\adjustbox{width=\linewidth}{
\begin{tabular}{cc|cccccc}
    \toprule
    $\Tc$ & $\Rc$ & PSNR $\uparrow$ & SSIM $\uparrow$ & LPIPS $\downarrow$ & CSIM $\uparrow$ & AED $\downarrow$ & APD $\downarrow$ \\
    \midrule
        \multirow{2}{*}{SF} & $-$ & 19.9 & 0.714 & 0.239 & 0.861 & 0.171 & 0.024 \\
        & SF & 20.0 & 0.722 & 0.237 & 0.872 & 0.165 & 0.023 \\
    \midrule
        \multirow{2}{*}{MLS} & $-$ & 20.1 & 0.718 & 0.230 & 0.879 & 0.157 & \textbf{0.015} \\
        & MLS & \textbf{20.3} & \textbf{0.730} & \textbf{0.226} & \textbf{0.896} & \textbf{0.148} & \textbf{0.015} \\
    \bottomrule
\end{tabular}
}
\caption{\textbf{Ablation study for deformation fields and illumination awareness on VFHQ-Test self-reenactment at resolution $512^2$.} $\Tc$ and $\Rc$ denote deformation fields for feature warping and normal rotation, respectively. No $\Rc$ means illumination unaware.}
\vspace{-1.0em}
\label{tab:ablation_mls}
\end{table}

\minisection{Impact of deformation fields.}
Rows 2 and 4 of~\cref{tab:ablation_mls} investigate the effect of using different deformation fields. The results show that using the MLS-based deformation field effectively enhances synthesis quality and motion control accuracy compared to its SF-based counterparts. These improvements highlight the effectiveness of MLS in maintaining smooth and consistent deformations.

\minisection{Impact of illumination awareness.}
As shown in every two rows of~\cref{tab:ablation_mls} (1\supst vs. 2\supnd and 3\suprd vs. 4\supth), removing the rotation field applied to canonical normals causes a noticeable decline in model performance. Without this rotation field, the normal directions remain fixed, even as pose and expression change, leaving the model illumination unaware and shadings stuck to the face of animated portraits.

\begin{table}[t]
\centering
\footnotesize
\setlength{\tabcolsep}{0.4em}
\adjustbox{width=\linewidth}{
\begin{tabular}{ccc|cccccc}
    \toprule
    Synthetic & Real & Regularization & CSIM $\uparrow$ & AED $\downarrow$ & APD $\downarrow$ & FID $\downarrow$ \\
    \midrule
        \cmark & \xmark & \xmark & 0.509 & 0.384 & 0.030 & 58.8 \\
        \xmark & \cmark & \xmark & 0.695 & 0.395 & \textbf{0.024} & 41.7 \\
        \cmark & \cmark & \xmark & 0.707 & 0.396 & 0.025 & 39.8 \\
        \cmark & \cmark & $\norm{\Sv(\Iv_\text{app})-\Sv(\Iv_\text{lit})}_1$ & \textbf{0.721} & 0.372 & 0.027 & 39.5 \\
        \cmark & \cmark & Random $\Iv_\text{lit}$ & 0.713 & \textbf{0.370} & \textbf{0.024} & \textbf{38.3} \\
    \bottomrule
\end{tabular}
}
\caption{\textbf{Ablation study for data and regularization schemes on VFHQ-Test cross-reenactment at resolution $512^2$.} Real data refers to VFHQ-Train. $\norm{\Sv(\Iv_\text{app})-\Sv(\Iv_\text{lit})}_1$: L1 loss regularizing the difference between lightmaps estimated from the appearance source and those from the lighting source. Random $\Iv_\text{lit}$: randomly choosing a lighting source from the same video clip for real data.}
\vspace{-1.0em}
\label{tab:ablation_data}
\end{table}

\minisection{Impact of data sources.} Rows 1-3 of~\cref{tab:ablation_data} validate the effectiveness of using both real and synthetic data in training \fullmodel. Synthetic data alone (row 1) enables motion transfer but lacks fidelity due to a distribution gap between synthetic subjects and real-world portraits. Real data alone (row 2) improves realism but leads to an unconstrained lightmap estimator in the end-to-end training stage, resulting in degraded performance. Combining both sources (row 3) achieves optimal results.

\minisection{Impact of regularization on real data.} Row 1-3 of~\cref{tab:ablation_data} follow standard reenactment training, where the appearance source $\Iv_\text{app}$ and the motion source $\Iv_\text{mot}$ are sampled from the same video, with $\Iv_\text{lit}=\Iv_\text{mot}$. Since $\Iv_\text{mot}$ is also the editing target, there is limited penalty for the lightmap estimator to misinterpret albedo as shading, e.g., beards as shadows, resulting in suboptimal performance. To tackle this, we test two types of regularization: enforcing consistency between lightmaps estimated from the appearance and lighting sources (row 4) and randomly sampling the lighting source within the same video clip (row 5). Both learn a more robust lightmap estimator that better decouples albedo from shading, enhancing fidelity for editing results. We choose row 5 as our final scheme for the best performance.

\subsection{Qualitative Analysis}

\begin{figure*}[t]
    \begin{center}
    \includegraphics[width=\linewidth]{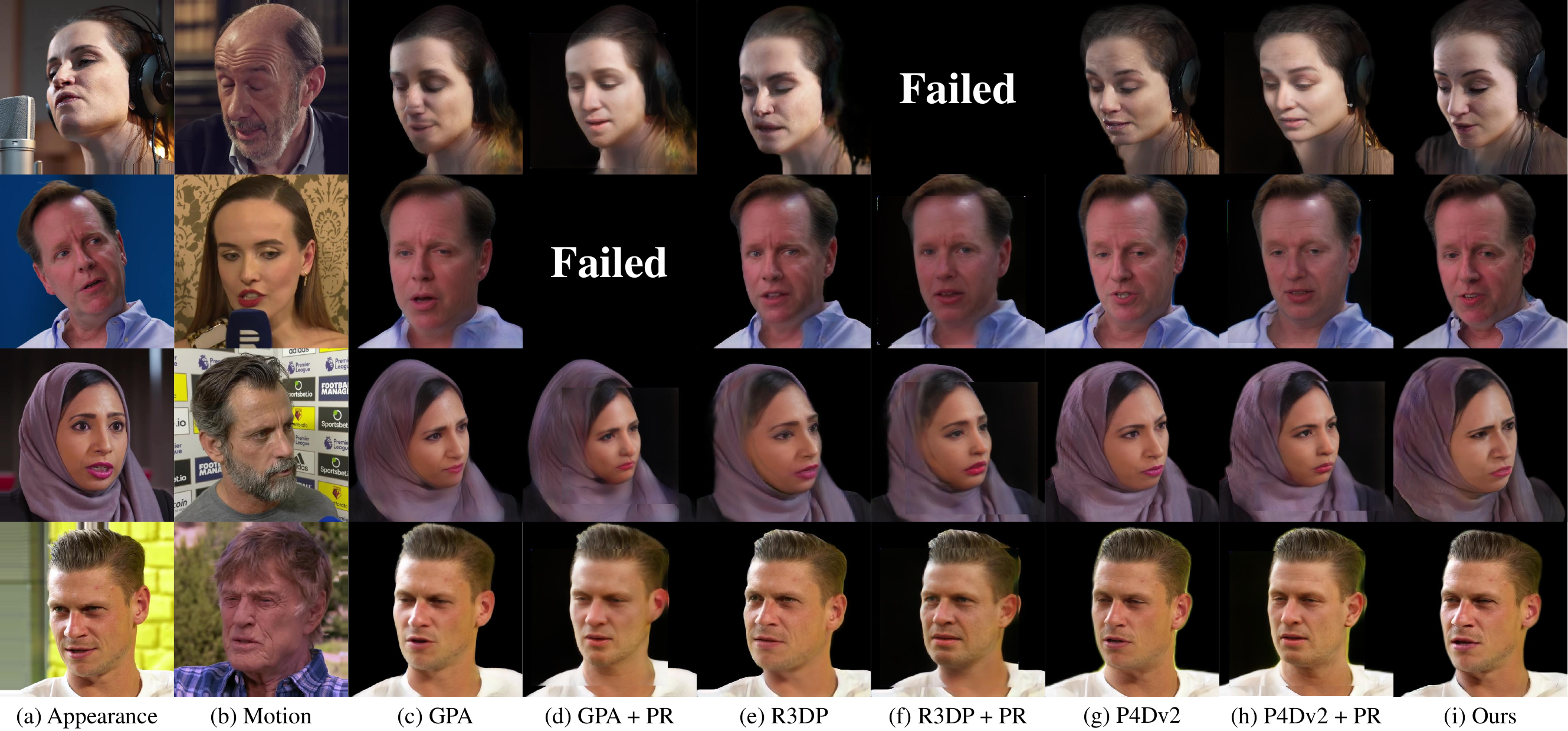}
    \vspace{-1.8em}
    \caption{\textbf{Qualitative comparison of cross-reenactment on VFHQ-Test.} For GPAvatar~\cite{chu2024gpavatar} (c), Real3DPortrait~\cite{ye2024real3d} (e), and Portrait4D-v2~\cite{deng2024portrait4dv2} (g), we use appearance sources (a) and motion sources (b) as inputs. For additional relighting with PortraitRelighting~\cite{cai2024real}, we use (a) as the lighting condition. For our \fullmodel, we use (a) as both appearance and lighting sources, and (b) as the motion source.} 
    \label{fig:qualitative_cross}
    \end{center}
    \vspace{-1.6em}
\end{figure*}

\begin{figure*}[t]
    \begin{center}
    \includegraphics[width=\linewidth]{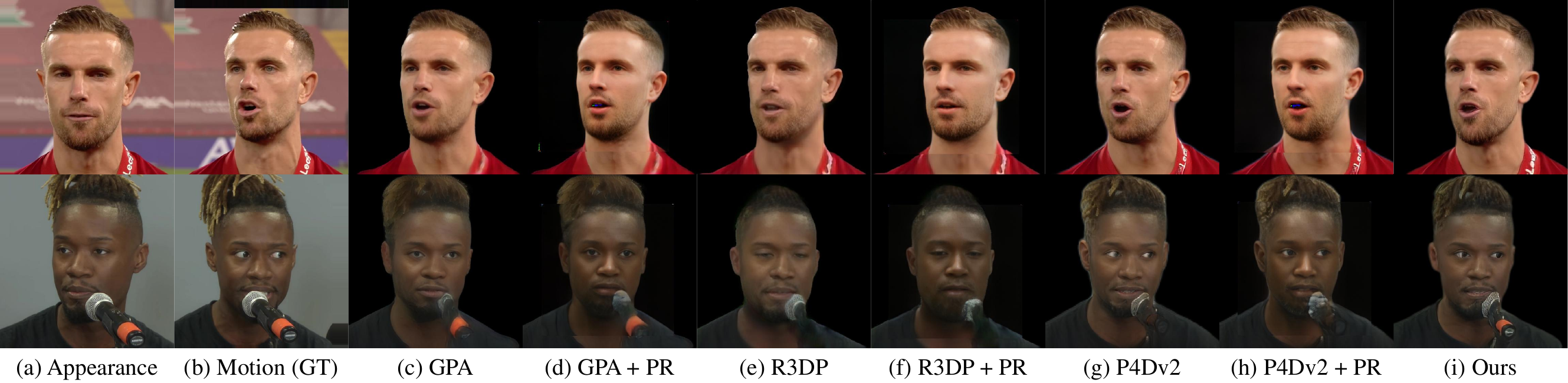}
    \vspace{-1.6em}
    \caption{\textbf{Qualitative comparison of self-reenactment on VFHQ-Test.} We employ the same input settings as in~\cref{fig:qualitative_cross}.}
    \label{fig:qualitative_self}
    \end{center}
    \vspace{-2.2em}
\end{figure*}

\begin{figure}[t]
    \begin{center}
    \includegraphics[width=\linewidth]{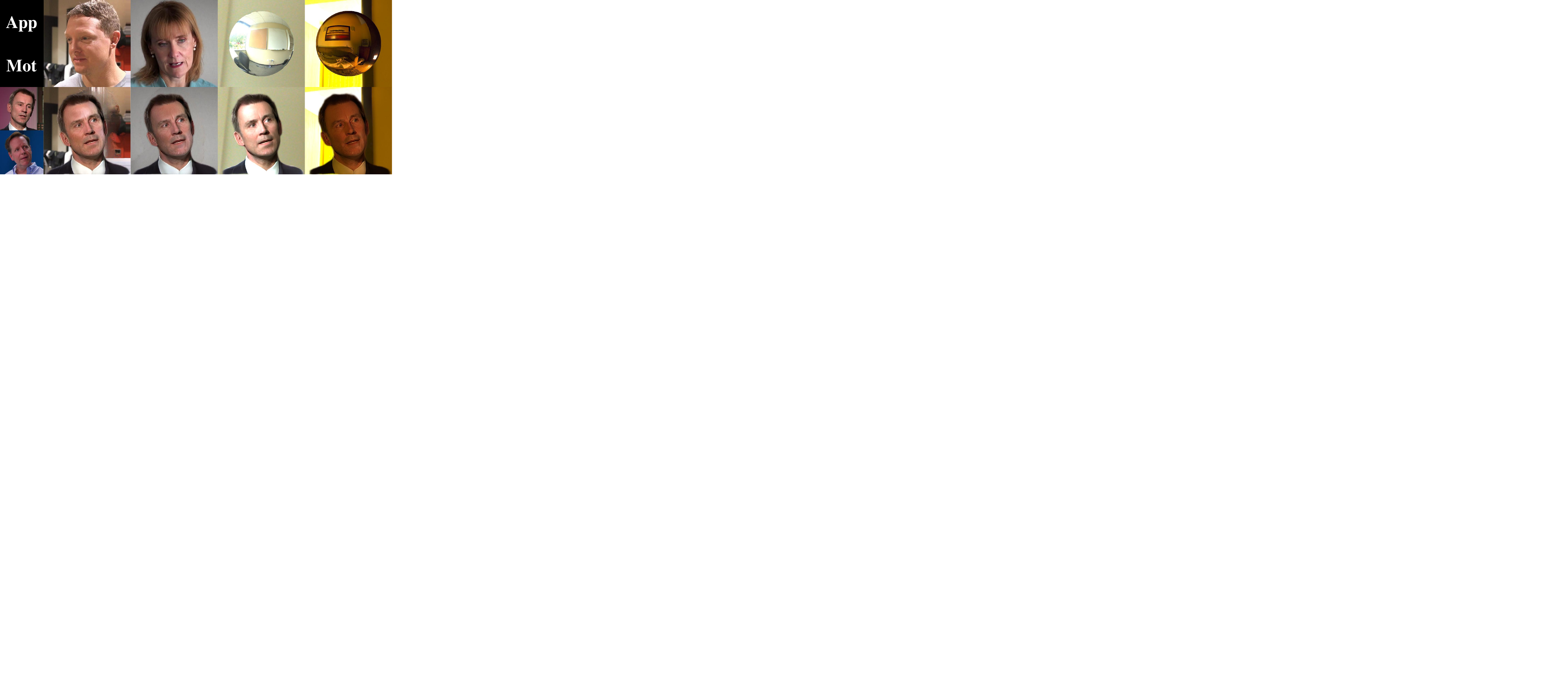}
    \vspace{-1.7em}
    \caption{\textbf{More applications.} \fullmodel enables flexible applications such as animatable portraits with background changing.}
    \label{fig:qualitative_relighting}
    \end{center}
    \vspace{-2.5em}
\end{figure}

\cref{fig:qualitative_cross} compares \fullmodel with other reenactment and reenactment-relighting pipelines for cross-enactment.
Face reenactment models tend to carry shading effects with portrait motion, leading to unrealistic results. Non-end-to-end solutions can fail if the reenactment step does not provide a reasonable portrait, compounding errors in subsequent relighting.
In contrast, our model achieves consistent, realistic results by handling both motion and lighting control.
\cref{fig:qualitative_self} compares \fullmodel with our baseline, Portrait4D-v2~\cite{deng2024portrait4dv2}, in terms of self-reenactment, where our method demonstrates better geometry and illumination control abilities.
Thanks to illumination awareness, our model synthesizes accurate movement of the light spot on the forehead.
Further, we explore the usage of \fullmodel in~\cref{fig:qualitative_relighting}.
Users can edit the appearance, motion, and lighting attributes of a portrait individually while not affecting others.
This facilitates applications such as background changing.
More visualizations can be found in \supp

\section{Conclusion}
We introduce \fullmodel, a geometry-and-illumination-aware portrait editing framework that synthesizes 3D portraits with given appearance, motion, and lighting sources.  With intrinsic decomposed neural radiance fields, it achieves precise lighting control using either a portrait image or an HDR environment map. The integration of a MLS-based deformation field further enhances the realism of the generated portraits. Experimental results show that \fullmodel provides superior performance and more flexible applications compared to existing methods.

\minisection{Limitation discussion and future work.}
The current formulation of \fullmodel does not account for visibility and self-occlusion, making it difficult to handle portraits with accessories, e.g., hats and glasses. However, as suggested in LumiGAN~\cite{deng2024lumigan}, voxel-wise visibilities can be calculated from the predicted density field and learned in a self-supervised manner. We leave this to future exploration.

\clearpage
{
    \small
    \bibliographystyle{ieeenat_fullname}
    \bibliography{main}

\begin{thebibliography}{84}
\providecommand{\natexlab}[1]{#1}
\providecommand{\url}[1]{\texttt{#1}}
\expandafter\ifx\csname urlstyle\endcsname\relax
  \providecommand{\doi}[1]{doi: #1}\else
  \providecommand{\doi}{doi: \begingroup \urlstyle{rm}\Url}\fi

\bibitem[Abdal et~al.(2021)Abdal, Zhu, Mitra, and Wonka]{abdal2021styleflow}
Rameen Abdal, Peihao Zhu, Niloy~J Mitra, and Peter Wonka.
\newblock Styleflow: Attribute-conditioned exploration of stylegan-generated
  images using conditional continuous normalizing flows.
\newblock \emph{ACM Transactions on Graphics (ToG)}, 40\penalty0 (3):\penalty0
  1--21, 2021.

\bibitem[Agarwal et~al.(2022)]{agarwal2022audio}
Madhav Agarwal et~al.
\newblock Audio-visual face reenactment.
\newblock \emph{arXiv preprint arXiv:2210.02755}, 2022.

\bibitem[Athar et~al.(2022)Athar, Xu, Sunkavalli, Shechtman, and
  Shu]{athar2022rignerf}
ShahRukh Athar, Zexiang Xu, Kalyan Sunkavalli, Eli Shechtman, and Zhixin Shu.
\newblock Rignerf: Fully controllable neural 3d portraits.
\newblock In \emph{Proceedings of the IEEE/CVF Conference on Computer Vision
  and Pattern Recognition (CVPR)}, pages 20364--20373, 2022.

\bibitem[Bergman et~al.(2022)Bergman, Kellnhofer, Yifan, Chan, Lindell, and
  Wetzstein]{bergman2022generative}
Alexander Bergman, Petr Kellnhofer, Wang Yifan, Eric Chan, David Lindell, and
  Gordon Wetzstein.
\newblock Generative neural articulated radiance fields.
\newblock \emph{Advances in Neural Information Processing Systems},
  35:\penalty0 19900--19916, 2022.

\bibitem[Blanz and Vetter(1999)]{blanz1999morphable}
Volker Blanz and Thomas Vetter.
\newblock A morphable model for the synthesis of 3d faces.
\newblock In \emph{Proceedings of the 26th annual conference on Computer
  graphics and interactive techniques}, pages 187--194. ACM
  Press/Addison-Wesley Publishing Co., 1999.

\bibitem[Cai et~al.(2024)Cai, Jiang, Chen, Lai, Fu, Shi, and Gao]{cai2024real}
Ziqi Cai, Kaiwen Jiang, Shu-Yu Chen, Yu-Kun Lai, Hongbo Fu, Boxin Shi, and Lin
  Gao.
\newblock Real-time 3d-aware portrait video relighting.
\newblock In \emph{Proceedings of the IEEE/CVF Conference on Computer Vision
  and Pattern Recognition}, pages 6221--6231, 2024.

\bibitem[Caselles et~al.(2023)Caselles, Ramon, Garcia, Giro-i Nieto,
  Moreno-Noguer, and Triginer]{caselles2023sira}
Pol Caselles, Eduard Ramon, Jaime Garcia, Xavier Giro-i Nieto, Francesc
  Moreno-Noguer, and Gil Triginer.
\newblock Sira: Relightable avatars from a single image.
\newblock In \emph{Proceedings of the IEEE/CVF winter conference on
  applications of computer vision}, pages 775--784, 2023.

\bibitem[Chan et~al.(2022)Chan, Lin, Chan, Nagano, Pan, De~Mello, Gallo,
  Guibas, Tremblay, Khamis, et~al.]{chan2022efficient}
Eric~R Chan, Connor~Z Lin, Matthew~A Chan, Koki Nagano, Boxiao Pan, Shalini
  De~Mello, Orazio Gallo, Leonidas~J Guibas, Jonathan Tremblay, Sameh Khamis,
  et~al.
\newblock Efficient geometry-aware 3d generative adversarial networks.
\newblock In \emph{Proceedings of the IEEE/CVF conference on computer vision
  and pattern recognition}, pages 16123--16133, 2022.

\bibitem[Chu and Harada(2024)]{chu2024generalizable}
Xuangeng Chu and Tatsuya Harada.
\newblock Generalizable and animatable gaussian head avatar.
\newblock \emph{arXiv preprint arXiv:2410.07971}, 2024.

\bibitem[Chu et~al.(2024)Chu, Li, Zeng, Yang, Lin, Liu, and
  Harada]{chu2024gpavatar}
Xuangeng Chu, Yu Li, Ailing Zeng, Tianyu Yang, Lijian Lin, Yunfei Liu, and
  Tatsuya Harada.
\newblock Gpavatar: Generalizable and precise head avatar from image (s).
\newblock \emph{arXiv preprint arXiv:2401.10215}, 2024.

\bibitem[Cook and Torrance(1982)]{cook1982reflectance}
Robert~L Cook and Kenneth~E. Torrance.
\newblock A reflectance model for computer graphics.
\newblock \emph{ACM Transactions on Graphics (ToG)}, 1\penalty0 (1):\penalty0
  7--24, 1982.

\bibitem[Debevec et~al.(2000)Debevec, Hawkins, Tchou, Duiker, Sarokin, and
  Sagar]{debevec2000acquiring}
Paul Debevec, Tim Hawkins, Chris Tchou, Haarm-Pieter Duiker, Westley Sarokin,
  and Mark Sagar.
\newblock Acquiring the reflectance field of a human face.
\newblock In \emph{Proceedings of the 27th annual conference on Computer
  graphics and interactive techniques}, pages 145--156, 2000.

\bibitem[Deng et~al.(2024{\natexlab{a}})Deng, Wang, and
  Wetzstein]{deng2024lumigan}
Boyang Deng, Yifan Wang, and Gordon Wetzstein.
\newblock Lumigan: Unconditional generation of relightable 3d human faces.
\newblock In \emph{2024 International Conference on 3D Vision (3DV)}, pages
  302--312. IEEE, 2024{\natexlab{a}}.

\bibitem[Deng et~al.(2019{\natexlab{a}})Deng, Guo, Xue, and
  Zafeiriou]{deng2019arcface}
Jiankang Deng, Jia Guo, Niannan Xue, and Stefanos Zafeiriou.
\newblock Arcface: Additive angular margin loss for deep face recognition.
\newblock In \emph{Proceedings of the IEEE/CVF conference on computer vision
  and pattern recognition}, pages 4690--4699, 2019{\natexlab{a}}.

\bibitem[Deng et~al.(2019{\natexlab{b}})Deng, Yang, Xu, Chen, Jia, and
  Tong]{deng2019accurate}
Yu Deng, Jiaolong Yang, Sicheng Xu, Dong Chen, Yunde Jia, and Xin Tong.
\newblock Accurate 3d face reconstruction with weakly-supervised learning: From
  single image to image set.
\newblock In \emph{Proceedings of the IEEE/CVF conference on computer vision
  and pattern recognition workshops}, pages 0--0, 2019{\natexlab{b}}.

\bibitem[Deng et~al.(2024{\natexlab{b}})Deng, Wang, Ren, Chen, and
  Wang]{deng2024portrait4d}
Yu Deng, Duomin Wang, Xiaohang Ren, Xingyu Chen, and Baoyuan Wang.
\newblock Portrait4d: Learning one-shot 4d head avatar synthesis using
  synthetic data.
\newblock In \emph{Proceedings of the IEEE/CVF Conference on Computer Vision
  and Pattern Recognition}, pages 7119--7130, 2024{\natexlab{b}}.

\bibitem[Deng et~al.(2024{\natexlab{c}})Deng, Wang, and
  Wang]{deng2024portrait4dv2}
Yu Deng, Duomin Wang, and Baoyuan Wang.
\newblock Portrait4d-v2: Pseudo multi-view data creates better 4d head
  synthesizer.
\newblock \emph{arXiv preprint arXiv:2403.13570}, 2024{\natexlab{c}}.

\bibitem[Fei et~al.(2023)Fei, Cheng, Zhu, Zheng, Li, Pan, and
  Shi]{fei2023split}
Fan Fei, Yean Cheng, Yongjie Zhu, Qian Zheng, Si Li, Gang Pan, and Boxin Shi.
\newblock Split: Single portrait lighting estimation via a tetrad of face
  intrinsics.
\newblock \emph{IEEE Transactions on Pattern Analysis and Machine
  Intelligence}, 2023.

\bibitem[Gafni et~al.(2021)Gafni, Thies, Zollh{\"o}fer, and
  Nie{\ss}ner]{gafni2021dynamic}
Guy Gafni, Justus Thies, Michael Zollh{\"o}fer, and Matthias Nie{\ss}ner.
\newblock Dynamic neural radiance fields for monocular 4d facial avatar
  reconstruction.
\newblock In \emph{Proceedings of the IEEE/CVF Conference on Computer Vision
  and Pattern Recognition (CVPR)}, pages 8649--8658, 2021.

\bibitem[Goodfellow et~al.(2014)Goodfellow, Pouget-Abadie, Mirza, Xu,
  Warde-Farley, Ozair, Courville, and Bengio]{goodfellow2014generative}
Ian Goodfellow, Jean Pouget-Abadie, Mehdi Mirza, Bing Xu, David Warde-Farley,
  Sherjil Ozair, Aaron Courville, and Yoshua Bengio.
\newblock Generative adversarial nets.
\newblock \emph{Advances in neural information processing systems}, 27, 2014.

\bibitem[{Google}(2021)]{google2021project}
{Google}.
\newblock Project starline.
\newblock \url{https://en.wikipedia.org/wiki/Project_Starline}, 2021.

\bibitem[Guo et~al.(2025)Guo, Xing, and Liu]{guo2025high}
Mingtao Guo, Guanyu Xing, and Yanli Liu.
\newblock High-fidelity relightable monocular portrait animation with
  lighting-controllable video diffusion model.
\newblock \emph{arXiv preprint arXiv:2502.19894}, 2025.

\bibitem[Heidrich and Seidel(1999)]{heidrich1999realistic}
Wolfgang Heidrich and Hans-Peter Seidel.
\newblock Realistic, hardware-accelerated shading and lighting.
\newblock In \emph{Proceedings of the 26th annual conference on Computer
  graphics and interactive techniques}, pages 171--178, 1999.

\bibitem[Heusel et~al.(2017)Heusel, Ramsauer, Unterthiner, Nessler, and
  Hochreiter]{heusel2017gans}
Martin Heusel, Hubert Ramsauer, Thomas Unterthiner, Bernhard Nessler, and Sepp
  Hochreiter.
\newblock Gans trained by a two time-scale update rule converge to a local nash
  equilibrium.
\newblock \emph{Advances in neural information processing systems}, 30, 2017.

\bibitem[Hong et~al.(2022)Hong, Zhang, Shen, and Xu]{hong2022depth}
Fa-Ting Hong, Longhao Zhang, Li Shen, and Dan Xu.
\newblock Depth-aware generative adversarial network for talking head video
  generation.
\newblock In \emph{Proceedings of the IEEE/CVF conference on computer vision
  and pattern recognition}, pages 3397--3406, 2022.

\bibitem[Hou et~al.(2021)Hou, Zhang, Sarkis, Bi, Tong, and Liu]{hou2021towards}
Andrew Hou, Ze Zhang, Michel Sarkis, Ning Bi, Yiying Tong, and Xiaoming Liu.
\newblock Towards high fidelity face relighting with realistic shadows.
\newblock In \emph{Proceedings of the IEEE/CVF conference on computer vision
  and pattern recognition}, pages 14719--14728, 2021.

\bibitem[Hou et~al.(2022)Hou, Sarkis, Bi, Tong, and Liu]{hou2022face}
Andrew Hou, Michel Sarkis, Ning Bi, Yiying Tong, and Xiaoming Liu.
\newblock Face relighting with geometrically consistent shadows.
\newblock In \emph{Proceedings of the IEEE/CVF conference on computer vision
  and pattern recognition}, pages 4217--4226, 2022.

\bibitem[Ji et~al.(2022)Ji, Yu, Guo, Liu, and Liu]{ji2022geometry}
Chaonan Ji, Tao Yu, Kaiwen Guo, Jingxin Liu, and Yebin Liu.
\newblock Geometry-aware single-image full-body human relighting.
\newblock In \emph{European Conference on Computer Vision}, pages 388--405.
  Springer, 2022.

\bibitem[Jiang et~al.(2023)Jiang, Chen, Fu, and Gao]{jiang2023nerffacelighting}
Kaiwen Jiang, Shu-Yu Chen, Hongbo Fu, and Lin Gao.
\newblock Nerffacelighting: Implicit and disentangled face lighting
  representation leveraging generative prior in neural radiance fields.
\newblock \emph{ACM Transactions on Graphics}, 42\penalty0 (3):\penalty0 1--18,
  2023.

\bibitem[Karras et~al.(2020)Karras, Laine, Aittala, Hellsten, Lehtinen, and
  Aila]{karras2020analyzing}
Tero Karras, Samuli Laine, Miika Aittala, Janne Hellsten, Jaakko Lehtinen, and
  Timo Aila.
\newblock Analyzing and improving the image quality of stylegan.
\newblock In \emph{Proceedings of the IEEE/CVF conference on computer vision
  and pattern recognition}, pages 8110--8119, 2020.

\bibitem[Kautz et~al.(2000)Kautz, V{\'a}zquez, Heidrich, and
  Seidel]{kautz2000unified}
Jan Kautz, Pere-Pau V{\'a}zquez, Wolfgang Heidrich, and Hans-Peter Seidel.
\newblock A unified approach to prefiltered environment maps.
\newblock In \emph{Rendering Techniques 2000: Proceedings of the Eurographics
  Workshop in Brno, Czech Republic, June 26--28, 2000 11}, pages 185--196.
  Springer, 2000.

\bibitem[Khakhulin et~al.(2022)Khakhulin, Sklyarova, Lempitsky, and
  Zakharov]{khakhulin2022realistic}
Taras Khakhulin, Vanessa Sklyarova, Victor Lempitsky, and Egor Zakharov.
\newblock Realistic one-shot mesh-based head avatars.
\newblock In \emph{European Conference on Computer Vision}, pages 345--362.
  Springer, 2022.

\bibitem[Khirodkar et~al.(2024)Khirodkar, Bagautdinov, Martinez, Zhaoen, James,
  Selednik, Anderson, and Saito]{khirodkar2024sapiens}
Rawal Khirodkar, Timur Bagautdinov, Julieta Martinez, Su Zhaoen, Austin James,
  Peter Selednik, Stuart Anderson, and Shunsuke Saito.
\newblock Sapiens: Foundation for human vision models.
\newblock \emph{arXiv preprint arXiv:2408.12569}, 2024.

\bibitem[Kim et~al.(2024)Kim, Jang, Yoon, Lee, Na, and Woo]{kim2024switchlight}
Hoon Kim, Minje Jang, Wonjun Yoon, Jisoo Lee, Donghyun Na, and Sanghyun Woo.
\newblock Switchlight: Co-design of physics-driven architecture and
  pre-training framework for human portrait relighting.
\newblock In \emph{Proceedings of the IEEE/CVF Conference on Computer Vision
  and Pattern Recognition}, pages 25096--25106, 2024.

\bibitem[LeCun et~al.(1998)LeCun, Bottou, Bengio, and
  Haffner]{lecun1998gradient}
Yann LeCun, L{\'e}on Bottou, Yoshua Bengio, and Patrick Haffner.
\newblock Gradient-based learning applied to document recognition.
\newblock \emph{Proceedings of the IEEE}, 86\penalty0 (11):\penalty0
  2278--2324, 1998.

\bibitem[Li et~al.(2023)Li, Zhang, Wang, Zhao, Wang, Chen, Zhang, Wang, Bo, and
  Li]{li2023one}
Weichuang Li, Longhao Zhang, Dong Wang, Bin Zhao, Zhigang Wang, Mulin Chen,
  Bang Zhang, Zhongjian Wang, Liefeng Bo, and Xuelong Li.
\newblock One-shot high-fidelity talking-head synthesis with deformable neural
  radiance field.
\newblock In \emph{Proceedings of the IEEE/CVF Conference on Computer Vision
  and Pattern Recognition}, pages 17969--17978, 2023.

\bibitem[Li et~al.(2024)Li, De~Mello, Liu, Nagano, Iqbal, and
  Kautz]{li2024generalizable}
Xueting Li, Shalini De~Mello, Sifei Liu, Koki Nagano, Umar Iqbal, and Jan
  Kautz.
\newblock Generalizable one-shot 3d neural head avatar.
\newblock \emph{Advances in Neural Information Processing Systems}, 36, 2024.

\bibitem[Lin et~al.(2022)Lin, Lindell, Chan, and Wetzstein]{lin20223d}
Connor~Z Lin, David~B Lindell, Eric~R Chan, and Gordon Wetzstein.
\newblock 3d gan inversion for controllable portrait image animation.
\newblock \emph{arXiv preprint arXiv:2203.13441}, 2022.

\bibitem[Lin et~al.(2024)Lin, Reddy, Berger, Sarkis, Porikli, and
  Bi]{lin2024edgerelight360}
Min-Hui Lin, Mahesh Reddy, Guillaume Berger, Michel Sarkis, Fatih Porikli, and
  Ning Bi.
\newblock Edgerelight360: Text-conditioned 360-degree hdr image generation for
  real-time on-device video portrait relighting.
\newblock In \emph{Proceedings of the IEEE/CVF Conference on Computer Vision
  and Pattern Recognition}, pages 831--840, 2024.

\bibitem[Ma et~al.(2023)Ma, Zhu, Qi, Lei, and Zhang]{ma2023otavatar}
Zhiyuan Ma, Xiangyu Zhu, Guo-Jun Qi, Zhen Lei, and Lei Zhang.
\newblock Otavatar: One-shot talking face avatar with controllable tri-plane
  rendering.
\newblock In \emph{Proceedings of the IEEE/CVF Conference on Computer Vision
  and Pattern Recognition}, pages 16901--16910, 2023.

\bibitem[Mei et~al.(2023)Mei, Zhang, Zhang, Zhang, Shu, Wang, Wei, Yan, Jung,
  and Patel]{mei2023lightpainter}
Yiqun Mei, He Zhang, Xuaner Zhang, Jianming Zhang, Zhixin Shu, Yilin Wang,
  Zijun Wei, Shi Yan, HyunJoon Jung, and Vishal~M Patel.
\newblock Lightpainter: interactive portrait relighting with freehand scribble.
\newblock In \emph{Proceedings of the IEEE/CVF Conference on Computer Vision
  and Pattern Recognition}, pages 195--205, 2023.

\bibitem[Mei et~al.(2024)Mei, Zeng, Zhang, Shu, Zhang, Bi, Zhang, Jung, and
  Patel]{mei2024holo}
Yiqun Mei, Yu Zeng, He Zhang, Zhixin Shu, Xuaner Zhang, Sai Bi, Jianming Zhang,
  HyunJoon Jung, and Vishal~M Patel.
\newblock Holo-relighting: Controllable volumetric portrait relighting from a
  single image.
\newblock In \emph{Proceedings of the IEEE/CVF Conference on Computer Vision
  and Pattern Recognition}, pages 4263--4273, 2024.

\bibitem[Mildenhall et~al.(2020)Mildenhall, Srinivasan, Tancik, Barron,
  Ramamoorthi, and Ng]{mildenhall2020nerf}
Ben Mildenhall, Pratul~P Srinivasan, Matthew Tancik, Jonathan~T Barron, Ravi
  Ramamoorthi, and Ren Ng.
\newblock Nerf: Representing scenes as neural radiance fields for view
  synthesis.
\newblock In \emph{European conference on computer vision}, pages 405--421.
  Springer, 2020.

\bibitem[Nestmeyer et~al.(2020)Nestmeyer, Lalonde, Matthews, and
  Lehrmann]{nestmeyer2020learning}
Thomas Nestmeyer, Jean-Fran{\c{c}}ois Lalonde, Iain Matthews, and Andreas
  Lehrmann.
\newblock Learning physics-guided face relighting under directional light.
\newblock In \emph{Proceedings of the IEEE/CVF Conference on Computer Vision
  and Pattern Recognition}, pages 5124--5133, 2020.

\bibitem[Pandey et~al.(2021)Pandey, Orts-Escolano, Legendre, Haene, Bouaziz,
  Rhemann, Debevec, and Fanello]{pandey2021total}
Rohit Pandey, Sergio Orts-Escolano, Chloe Legendre, Christian Haene, Sofien
  Bouaziz, Christoph Rhemann, Paul~E Debevec, and Sean~Ryan Fanello.
\newblock Total relighting: learning to relight portraits for background
  replacement.
\newblock \emph{ACM Trans. Graph.}, 40\penalty0 (4):\penalty0 43--1, 2021.

\bibitem[Papantoniou et~al.(2023)Papantoniou, Lattas, Moschoglou, and
  Zafeiriou]{papantoniou2023relightify}
Foivos~Paraperas Papantoniou, Alexandros Lattas, Stylianos Moschoglou, and
  Stefanos Zafeiriou.
\newblock Relightify: Relightable 3d faces from a single image via diffusion
  models.
\newblock In \emph{Proceedings of the IEEE/CVF International Conference on
  Computer Vision}, pages 8806--8817, 2023.

\bibitem[Park et~al.(2021{\natexlab{a}})Park, Sinha, Barron, Bouaziz, Goldman,
  Seitz, and Martin-Brualla]{park2021nerfies}
Keunhong Park, Utkarsh Sinha, Jonathan~T. Barron, Sofien Bouaziz, Dan~B.
  Goldman, Steven~M. Seitz, and Ricardo Martin-Brualla.
\newblock Nerfies: Deformable neural radiance fields.
\newblock In \emph{Proceedings of the IEEE/CVF International Conference on
  Computer Vision (ICCV)}, pages 5865--5874, 2021{\natexlab{a}}.

\bibitem[Park et~al.(2021{\natexlab{b}})Park, Sinha, Hedman, Barron, Bouaziz,
  Goldman, Martin-Brualla, and Seitz]{park2021hypernerf}
Keunhong Park, Utkarsh Sinha, Peter Hedman, Jonathan~T. Barron, Sofien Bouaziz,
  Dan~B. Goldman, Ricardo Martin-Brualla, and Steven~M. Seitz.
\newblock Hypernerf: A higher-dimensional representation for topologically
  varying neural radiance fields.
\newblock \emph{ACM Transactions on Graphics (TOG)}, 40\penalty0 (6):\penalty0
  1--12, 2021{\natexlab{b}}.

\bibitem[Phong(1998)]{phong1998illumination}
Bui~Tuong Phong.
\newblock Illumination for computer generated pictures.
\newblock In \emph{Seminal graphics: pioneering efforts that shaped the field},
  pages 95--101. 1998.

\bibitem[Ponglertnapakorn et~al.(2023)Ponglertnapakorn, Tritrong, and
  Suwajanakorn]{ponglertnapakorn2023difareli}
Puntawat Ponglertnapakorn, Nontawat Tritrong, and Supasorn Suwajanakorn.
\newblock Difareli: Diffusion face relighting.
\newblock In \emph{Proceedings of the IEEE/CVF International Conference on
  Computer Vision}, pages 22646--22657, 2023.

\bibitem[Qiu et~al.(2024)Qiu, Chen, Jiang, Zhou, Fan, Yang, Wu, and
  Liu]{qiu2024relitalk}
Haonan Qiu, Zhaoxi Chen, Yuming Jiang, Hang Zhou, Xiangyu Fan, Lei Yang, Wayne
  Wu, and Ziwei Liu.
\newblock Relitalk: Relightable talking portrait generation from a single
  video.
\newblock \emph{International Journal of Computer Vision}, pages 1--16, 2024.

\bibitem[Ramamoorthi and Hanrahan(2001)]{ramamoorthi2001efficient}
Ravi Ramamoorthi and Pat Hanrahan.
\newblock An efficient representation for irradiance environment maps.
\newblock In \emph{Proceedings of the 28th annual conference on Computer
  graphics and interactive techniques}, pages 497--500, 2001.

\bibitem[Ranjan et~al.(2023)Ranjan, Yi, Chang, and Tuzel]{ranjan2023facelit}
Anurag Ranjan, Kwang~Moo Yi, Jen-Hao~Rick Chang, and Oncel Tuzel.
\newblock Facelit: Neural 3d relightable faces.
\newblock In \emph{Proceedings of the IEEE/CVF Conference on Computer Vision
  and Pattern Recognition}, pages 8619--8628, 2023.

\bibitem[Rao et~al.(2024)Rao, Fox, Meka, BR, Zhan, Weyrich, Bickel, Pfister,
  Matusik, Elgharib, et~al.]{rao2024lite2relight}
Pramod Rao, Gereon Fox, Abhimitra Meka, Mallikarjun BR, Fangneng Zhan, Tim
  Weyrich, Bernd Bickel, Hanspeter Pfister, Wojciech Matusik, Mohamed Elgharib,
  et~al.
\newblock Lite2relight: 3d-aware single image portrait relighting.
\newblock In \emph{ACM SIGGRAPH 2024 Conference Papers}, pages 1--12, 2024.

\bibitem[Ren et~al.(2024)Ren, Xiong, Yoon, Shu, Zhang, Jung, Gerig, and
  Zhang]{ren2024relightful}
Mengwei Ren, Wei Xiong, Jae~Shin Yoon, Zhixin Shu, Jianming Zhang, HyunJoon
  Jung, Guido Gerig, and He Zhang.
\newblock Relightful harmonization: Lighting-aware portrait background
  replacement.
\newblock In \emph{Proceedings of the IEEE/CVF Conference on Computer Vision
  and Pattern Recognition}, pages 6452--6462, 2024.

\bibitem[Ren et~al.(2021)Ren, Liu, Jiang, Liang, and Lin]{ren2021adaptive}
Yang Ren, Jie Liu, Xinwei Jiang, Xiaodan Liang, and Liang Lin.
\newblock Adaptive perturbation learning for unsupervised disentangling of
  appearance and motion.
\newblock In \emph{Proceedings of the IEEE/CVF Conference on Computer Vision
  and Pattern Recognition (CVPR)}, pages 14318--14327, 2021.

\bibitem[Saito et~al.(2024)Saito, Schwartz, Simon, Li, and
  Nam]{saito2024relightable}
Shunsuke Saito, Gabriel Schwartz, Tomas Simon, Junxuan Li, and Giljoo Nam.
\newblock Relightable gaussian codec avatars.
\newblock In \emph{Proceedings of the IEEE/CVF Conference on Computer Vision
  and Pattern Recognition}, pages 130--141, 2024.

\bibitem[Schaefer et~al.(2006)Schaefer, McPhail, and Warren]{schaefer2006image}
Scott Schaefer, Travis McPhail, and Joe Warren.
\newblock Image deformation using moving least squares.
\newblock In \emph{ACM SIGGRAPH 2006 Papers}, pages 533--540. 2006.

\bibitem[Siarohin et~al.(2019)Siarohin, Lathuili{\`e}re, Tulyakov, Ricci, and
  Sebe]{siarohin2019first}
Aliaksandr Siarohin, St{\'e}phane Lathuili{\`e}re, Sergey Tulyakov, Elisa
  Ricci, and Nicu Sebe.
\newblock First order motion model for image animation.
\newblock \emph{Advances in neural information processing systems}, 32, 2019.

\bibitem[Sun et~al.(2019)Sun, Barron, Tsai, Xu, Yu, Fyffe, Rhemann, Busch,
  Debevec, and Ramamoorthi]{sun2019single}
Tiancheng Sun, Jonathan~T Barron, Yun-Ta Tsai, Zexiang Xu, Xueming Yu, Graham
  Fyffe, Christoph Rhemann, Jay Busch, Paul Debevec, and Ravi Ramamoorthi.
\newblock Single image portrait relighting.
\newblock \emph{ACM Transactions on Graphics (TOG)}, 38\penalty0 (4):\penalty0
  1--12, 2019.

\bibitem[Sun et~al.(2021)Sun, Lin, Bi, Xu, and Ramamoorthi]{sun2021nelf}
Tiancheng Sun, Kai-En Lin, Sai Bi, Zexiang Xu, and Ravi Ramamoorthi.
\newblock Nelf: Neural light-transport field for portrait view synthesis and
  relighting.
\newblock \emph{arXiv preprint arXiv:2107.12351}, 2021.

\bibitem[Suvorov et~al.(2022)Suvorov, Logacheva, Mashikhin, Remizova, Ashukha,
  Silvestrov, Kong, Goka, Park, and Lempitsky]{suvorov2022resolution}
Roman Suvorov, Elizaveta Logacheva, Anton Mashikhin, Anastasia Remizova,
  Arsenii Ashukha, Aleksei Silvestrov, Naejin Kong, Harshith Goka, Kiwoong
  Park, and Victor Lempitsky.
\newblock Resolution-robust large mask inpainting with fourier convolutions.
\newblock In \emph{Proceedings of the IEEE/CVF winter conference on
  applications of computer vision}, pages 2149--2159, 2022.

\bibitem[Tang et~al.(2023)Tang, Zhang, Yang, Zhang, Chen, Ma, and
  Wen]{tang20233dfaceshop}
Junshu Tang, Bo Zhang, Binxin Yang, Ting Zhang, Dong Chen, Lizhuang Ma, and
  Fang Wen.
\newblock {3DFaceShop}: Explicitly controllable 3d-aware portrait generation.
\newblock \emph{IEEE Transactions on Visualization and Computer Graphics},
  2023.

\bibitem[Thies et~al.(2019)Thies, Zollh{\"o}fer, Stamminger, Theobalt, and
  Nie{\ss}ner]{thies2019neural}
Justus Thies, Michael Zollh{\"o}fer, Marc Stamminger, Christian Theobalt, and
  Matthias Nie{\ss}ner.
\newblock Neural voice puppetry: Audio-driven facial reenactment.
\newblock In \emph{Proceedings of the IEEE/CVF Conference on Computer Vision
  and Pattern Recognition}, pages 9382--9391, 2019.

\bibitem[Tretschk et~al.(2021)Tretschk, Tewari, Golyanik, Zollh{\"o}fer, and
  Theobalt]{tretschk2021nonrigid}
Edgar Tretschk, Ayush Tewari, Vladislav Golyanik, Michael Zollh{\"o}fer, and
  Christian Theobalt.
\newblock Non-rigid neural radiance fields: Reconstruction and novel view
  synthesis of a deforming scene from monocular video.
\newblock In \emph{Proceedings of the IEEE/CVF International Conference on
  Computer Vision (ICCV)}, pages 12959--12970, 2021.

\bibitem[Trevithick et~al.(2023)Trevithick, Chan, Stengel, Chan, Liu, Yu,
  Khamis, Ramamoorthi, and Nagano]{trevithick2023real}
Alex Trevithick, Matthew Chan, Michael Stengel, Eric Chan, Chao Liu, Zhiding
  Yu, Sameh Khamis, Ravi Ramamoorthi, and Koki Nagano.
\newblock Real-time radiance fields for single-image portrait view synthesis.
\newblock 2023.

\bibitem[Wang et~al.(2023)Wang, Deng, Yin, Shum, and Wang]{wang2023progressive}
Duomin Wang, Yu Deng, Zixin Yin, Heung-Yeung Shum, and Baoyuan Wang.
\newblock Progressive disentangled representation learning for fine-grained
  controllable talking head synthesis.
\newblock In \emph{Proceedings of the IEEE/CVF Conference on Computer Vision
  and Pattern Recognition}, pages 17979--17989, 2023.

\bibitem[Wang et~al.(2020{\natexlab{a}})Wang, Mallya, and Liu]{wang2020one}
Ting-Chun Wang, Arun Mallya, and Ming-Yu Liu.
\newblock One-shot free-view neural talking-head synthesis for video
  conferencing.
\newblock \emph{arXiv preprint arXiv:2011.15126}, 2020{\natexlab{a}}.

\bibitem[Wang et~al.(2020{\natexlab{b}})Wang, Yu, Lu, Wang, Qian, and
  Xu]{wang2020single}
Zhibo Wang, Xin Yu, Ming Lu, Quan Wang, Chen Qian, and Feng Xu.
\newblock Single image portrait relighting via explicit multiple reflectance
  channel modeling.
\newblock \emph{ACM Transactions on Graphics (ToG)}, 39\penalty0 (6):\penalty0
  1--13, 2020{\natexlab{b}}.

\bibitem[Wenger et~al.(2005)Wenger, Gardner, Tchou, Unger, Hawkins, and
  Debevec]{wenger2005performance}
Andreas Wenger, Andrew Gardner, Chris Tchou, Jonas Unger, Tim Hawkins, and Paul
  Debevec.
\newblock Performance relighting and reflectance transformation with
  time-multiplexed illumination.
\newblock \emph{ACM Transactions on Graphics (TOG)}, 24\penalty0 (3):\penalty0
  756--764, 2005.

\bibitem[Xie et~al.(2022)Xie, Wang, Zhang, Dong, and Shan]{xie2022vfhq}
Liangbin Xie, Xintao Wang, Honglun Zhang, Chao Dong, and Ying Shan.
\newblock Vfhq: A high-quality dataset and benchmark for video face
  super-resolution.
\newblock In \emph{Proceedings of the IEEE/CVF Conference on Computer Vision
  and Pattern Recognition}, pages 657--666, 2022.

\bibitem[Xu et~al.(2023{\natexlab{a}})Xu, Song, Jiang, Zhang, Shi, Liu, Ma,
  Feng, and Luo]{xu2023omniavatar}
Hongyi Xu, Guoxian Song, Zihang Jiang, Jianfeng Zhang, Yichun Shi, Jing Liu,
  Wanchun Ma, Jiashi Feng, and Linjie Luo.
\newblock Omniavatar: Geometry-guided controllable 3d head synthesis.
\newblock In \emph{Proceedings of the IEEE/CVF Conference on Computer Vision
  and Pattern Recognition (CVPR)}, pages 12814--12824, 2023{\natexlab{a}}.

\bibitem[Xu et~al.(2023{\natexlab{b}})Xu, Zhang, Liew, Zhang, Bai, Feng, and
  Shou]{xu2023pv3d}
Zhongcong Xu, Jianfeng Zhang, Junhao Liew, Wenqing Zhang, Song Bai, Jiashi
  Feng, and Mike~Zheng Shou.
\newblock {PV3D}: A 3d generative model for portrait video generation.
\newblock In \emph{Proceedings of the Tenth International Conference on
  Learning Representations (ICLR)}, 2023{\natexlab{b}}.

\bibitem[Ye et~al.(2024)Ye, Zhong, Ren, Yang, Li, Huang, Jiang, He, Huang, Liu,
  et~al.]{ye2024real3d}
Zhenhui Ye, Tianyun Zhong, Yi Ren, Jiaqi Yang, Weichuang Li, Jiawei Huang,
  Ziyue Jiang, Jinzheng He, Rongjie Huang, Jinglin Liu, et~al.
\newblock Real3d-portrait: One-shot realistic 3d talking portrait synthesis.
\newblock \emph{arXiv preprint arXiv:2401.08503}, 2024.

\bibitem[Yeh et~al.(2022)Yeh, Nagano, Khamis, Kautz, Liu, and
  Wang]{yeh2022learning}
Yu-Ying Yeh, Koki Nagano, Sameh Khamis, Jan Kautz, Ming-Yu Liu, and Ting-Chun
  Wang.
\newblock Learning to relight portrait images via a virtual light stage and
  synthetic-to-real adaptation.
\newblock \emph{ACM Transactions on Graphics (TOG)}, 41\penalty0 (6):\penalty0
  1--21, 2022.

\bibitem[Yu et~al.(2023{\natexlab{a}})Yu, Fan, Zhang, Wang, Yin, Bai, Cao,
  Shan, Wu, Sun, et~al.]{yu2023nofa}
Wangbo Yu, Yanbo Fan, Yong Zhang, Xuan Wang, Fei Yin, Yunpeng Bai, Yan-Pei Cao,
  Ying Shan, Yang Wu, Zhongqian Sun, et~al.
\newblock Nofa: Nerf-based one-shot facial avatar reconstruction.
\newblock In \emph{ACM SIGGRAPH 2023 Conference Proceedings}, pages 1--12,
  2023{\natexlab{a}}.

\bibitem[Yu et~al.(2023{\natexlab{b}})]{yu2023dinet}
Zhenyu Yu et~al.
\newblock Dinet: Deformation inpainting network for realistic face visually
  dubbing on high resolution video.
\newblock \emph{arXiv preprint arXiv:2304.10212}, 2023{\natexlab{b}}.

\bibitem[Zakharov et~al.(2019)Zakharov, Shysheya, Burkov, and
  Lempitsky]{zakharov2019few}
Egor Zakharov, Aliaksandra Shysheya, Egor Burkov, and Victor Lempitsky.
\newblock Few-shot adversarial learning of realistic neural talking head
  models.
\newblock In \emph{Proceedings of the IEEE/CVF International Conference on
  Computer Vision}, pages 9459--9468, 2019.

\bibitem[Zhang et~al.(2021{\natexlab{a}})Zhang, Zhang, Wu, Yu, and
  Xu]{zhang2021neural}
Longwen Zhang, Qixuan Zhang, Minye Wu, Jingyi Yu, and Lan Xu.
\newblock Neural video portrait relighting in real-time via consistency
  modeling.
\newblock In \emph{Proceedings of the IEEE/CVF international conference on
  computer vision}, pages 802--812, 2021{\natexlab{a}}.

\bibitem[Zhang et~al.(2018)Zhang, Isola, Efros, Shechtman, and
  Wang]{zhang2018unreasonable}
Richard Zhang, Phillip Isola, Alexei~A Efros, Eli Shechtman, and Oliver Wang.
\newblock The unreasonable effectiveness of deep features as a perceptual
  metric.
\newblock In \emph{Proceedings of the IEEE conference on computer vision and
  pattern recognition}, pages 586--595, 2018.

\bibitem[Zhang et~al.(2021{\natexlab{b}})Zhang, Li, Ding, and
  Fan]{zhang2021flow}
Zhimeng Zhang, Lincheng Li, Yu Ding, and Changjie Fan.
\newblock Flow-guided one-shot talking face generation with a high-resolution
  audio-visual dataset.
\newblock In \emph{Proceedings of the IEEE/CVF conference on computer vision
  and pattern recognition}, pages 3661--3670, 2021{\natexlab{b}}.

\bibitem[Zhou et~al.(2019)Zhou, Hadap, Sunkavalli, and Jacobs]{zhou2019deep}
Hao Zhou, Sunil Hadap, Kalyan Sunkavalli, and David~W Jacobs.
\newblock Deep single-image portrait relighting.
\newblock In \emph{Proceedings of the IEEE/CVF international conference on
  computer vision}, pages 7194--7202, 2019.

\bibitem[Zhu and Gortler(2007)]{zhu20073d}
Yuanchen Zhu and Steven~J Gortler.
\newblock 3d deformation using moving least squares.
\newblock 2007.

\bibitem[Zhuang et~al.(2022)Zhuang, Ma, Koyejo, and
  Schwing]{zhuang2022controllable}
Peiye Zhuang, Liqian Ma, Sanmi Koyejo, and Alexander Schwing.
\newblock Controllable radiance fields for dynamic face synthesis.
\newblock In \emph{Proceedings of the 2022 International Conference on 3D
  Vision (3DV)}, 2022.

\end{thebibliography}
}

\clearpage
\setcounter{page}{1}
\setcounter{section}{0}
\renewcommand{\thesection}{\Alph{section}}
\maketitlesupplementary

\section{Training Scheme and Objective Functions}
\label{sec:objective}
We begin by pre-training our lightmap estimator using synthetic data with ground truths. The objective function is
\begin{align}
    \Lc_\text{pre} = \Lc_{\Sv} + \Lc_{\Nv},
\end{align}
where $\Lc_{\Sv}$ is an L1 loss comparing the predicted and ground truth lightmaps $\Sv_\text{d},\{\Sv_\text{s}(n)\}$, $\Lc_{\Nv}$ is a cosine similarity loss between predicted and ground truth normal $\Nv$. After pre-training, we detach the normal decoder $\Ec_\text{nor}$ and integrate the rest of the lightmap estimator with \fullmodel. The entire network is then trained end-to-end using both real and synthetic data. During training, we randomly sample appearance source $\Iv_\text{app}$ and motion source $\Iv_\text{mot}$ of one subject, with the editing target $\Iv^*$ equal to the motion source $\Iv_\text{mot}$. As for the lighting source, we use the HDR environment map of $\Iv_\text{mot}$ for synthetic data and another random frame from the same video clip for real data. In this phase, our reconstruction objective is
\begin{align}
    \Lc_\text{rec} &= \Lc_1 + \Lc_\text{LPIPS} + \Lc_\text{id} + \Lc_\text{seg} + \Lc_{\av} + \Lc_{\nv} + \Lc_{\Sv},
\end{align}
where $\Lc_1$, $\Lc_\text{LPIPS}$, $\Lc_\text{id}$ are pixel-wise L1, perceptual difference~\cite{zhang2018unreasonable}, and negative cosine similarity of face recognition features~\cite{deng2019arcface} between the editing result $\hat{\Iv}$ and the target $\Iv^*$, $\Lc_\text{seg}$ and $\Lc_{\av}$ are L1 losses for the rendered foreground mask and albedo, $\Lc_{\nv}$ is a cosine similarity loss for the rendered normal, and $\Lc_{\Sv}$ is the L1 loss for estimated lightmaps. Note that $\Lc_{\av}$ and $\Lc_{\Sv}$ are used only for synthetic data, while $\Lc_{\nv}$ is also applied to real data with Sapiens~\cite{khirodkar2024sapiens} pseudo ground truths. Further, we introduce regularization
\begin{align}
    \Lc_\text{reg} &= \Rc_\text{TV} + \Rc_{\delta} + \Rc_{\nv},
\end{align}
where $\Rc_\text{TV}$ is the total variation loss to promote spatial smoothness, $\Rc_{\delta}$ is a L1 regularization for residual color $\delta\cv$ which constraints it from dominating the render, and
\begin{align}
    \Rc_{\nv} &= \norm{1 - \nv\cdot\left(-\frac{\nabla\sigma(\pv)}{\norm{\nabla\sigma(\pv)}_2}\right)}_1
\end{align}
regularizes normal $\nv$ to align with the unit negative gradient of density $\sigma$. We also apply an adversarial loss $\Lc_\text{adv}$ with a dual discriminator~\cite{chan2022efficient}. Finally, the training objective is
\begin{align}
    \Lc &= \Lc_\text{rec} + \Lc_\text{reg} + \Lc_\text{adv}.
\end{align}

\begin{figure}[t]
    \centering
    \includegraphics[width=\linewidth]{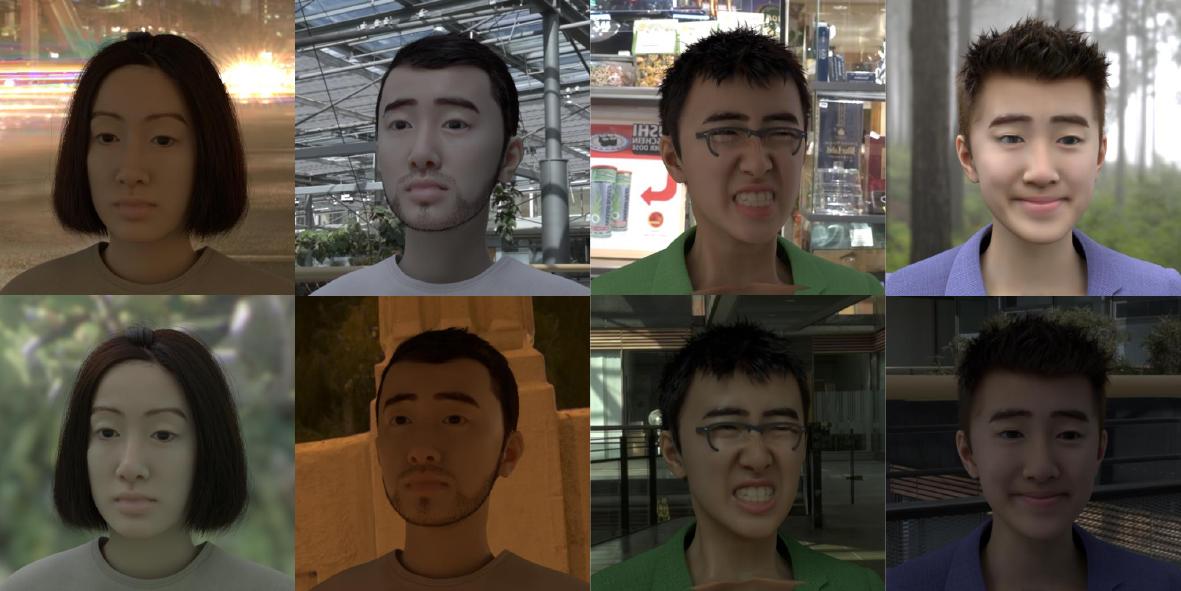}
    \caption{\textbf{Samples from one identity of the Lumos~\cite{yeh2022learning} dataset.} We refer to each column as a unique subject, since they have different appearances and accessories.}
    \label{fig:lumos_dataset}
    \vspace{-1.0em}
\end{figure}

\section{More Qualitative Results}
\begin{figure*}[t]
    \centering
    \includegraphics[width=\linewidth]{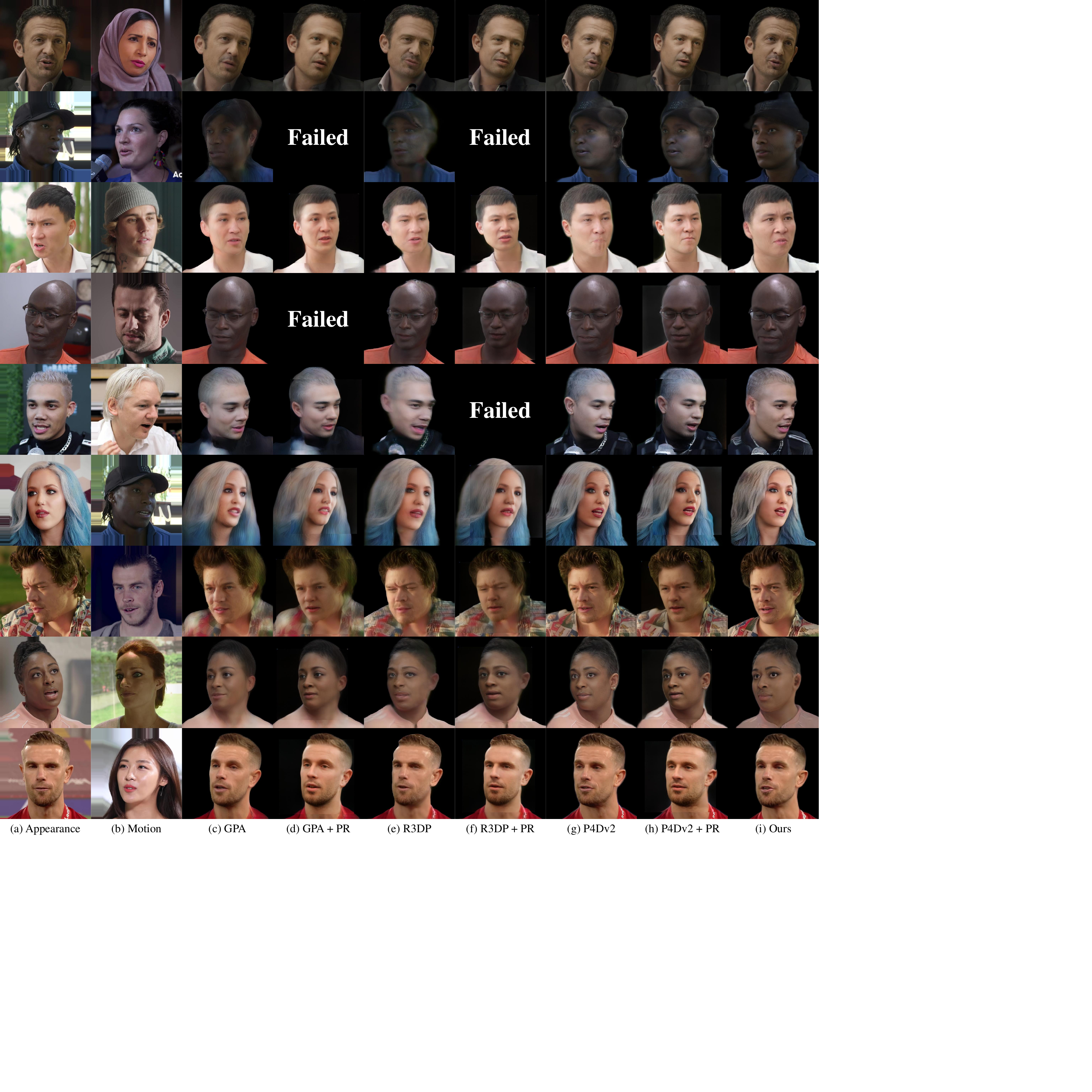}
    \caption{\textbf{Additional cross-reenactment results on the VFHQ dataset.}}
    \label{fig:additional_cross_vfhq}
\end{figure*}

\begin{figure*}[t]
    \centering
    \includegraphics[width=\linewidth]{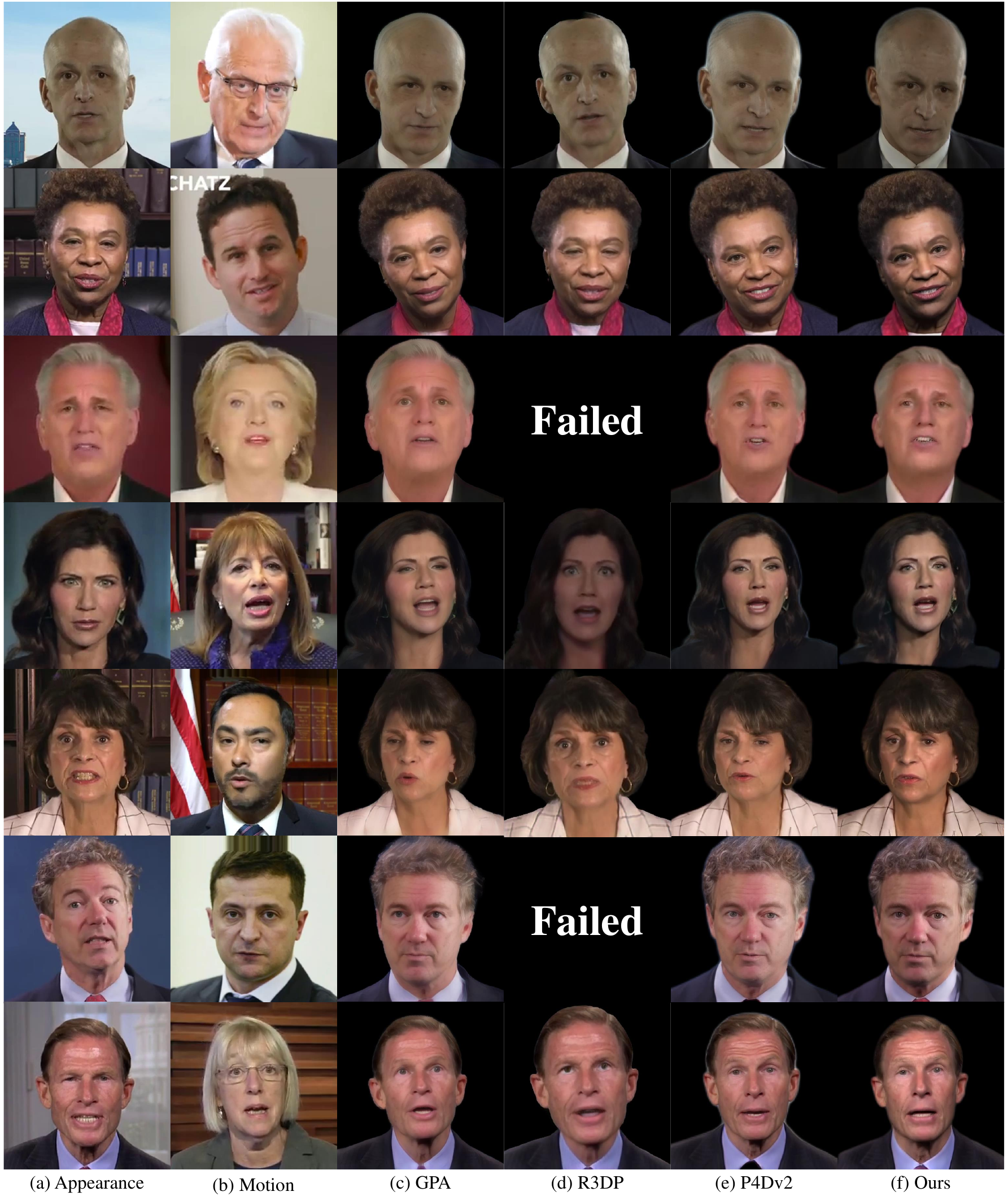}
    \caption{\textbf{Cross-reenactment results on the HTDF dataset.}}
    \label{fig:cross_hdtf_p1}
\end{figure*}

\begin{figure*}[t]
    \centering
    \includegraphics[width=0.97\linewidth]{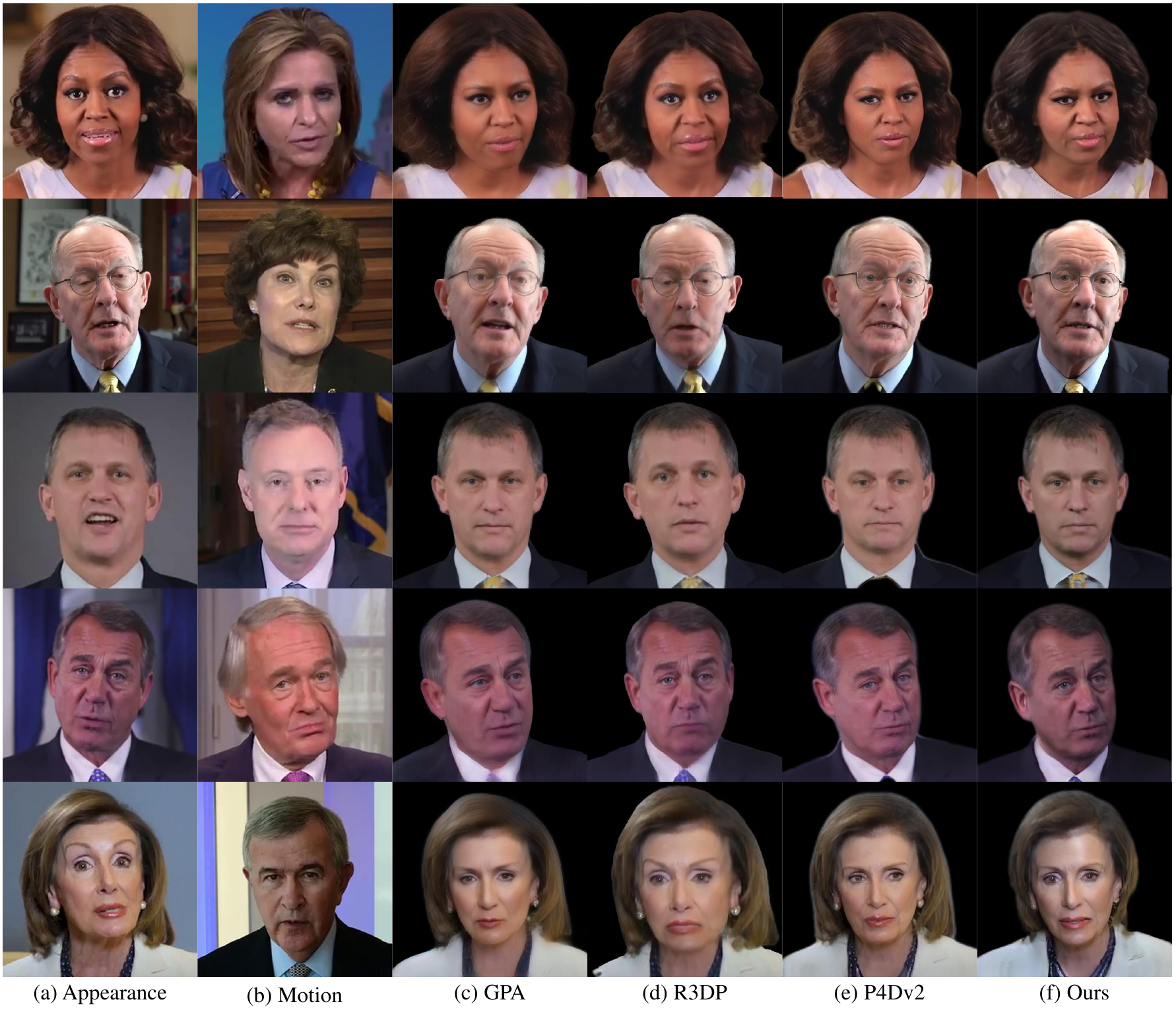}
    \caption{\textbf{Cross-reenactment results on the HTDF dataset (continued).}}
    \label{fig:cross_hdtf_p2}
\end{figure*}

\begin{figure*}[t]
    \centering
    \includegraphics[width=0.97\linewidth]{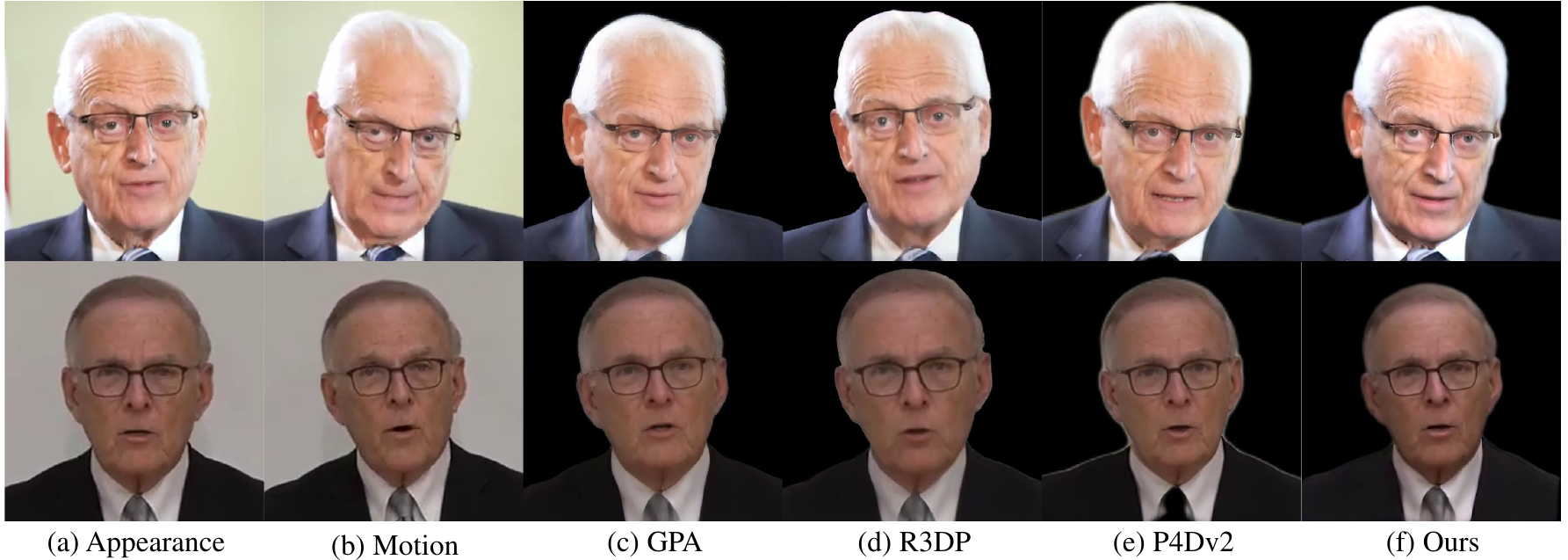}
    \caption{\textbf{Self-reenactment results on the HTDF dataset.}}
    \label{fig:self_hdtf}
\end{figure*}

\begin{figure*}[t]
    \centering
    \includegraphics[width=0.97\linewidth]{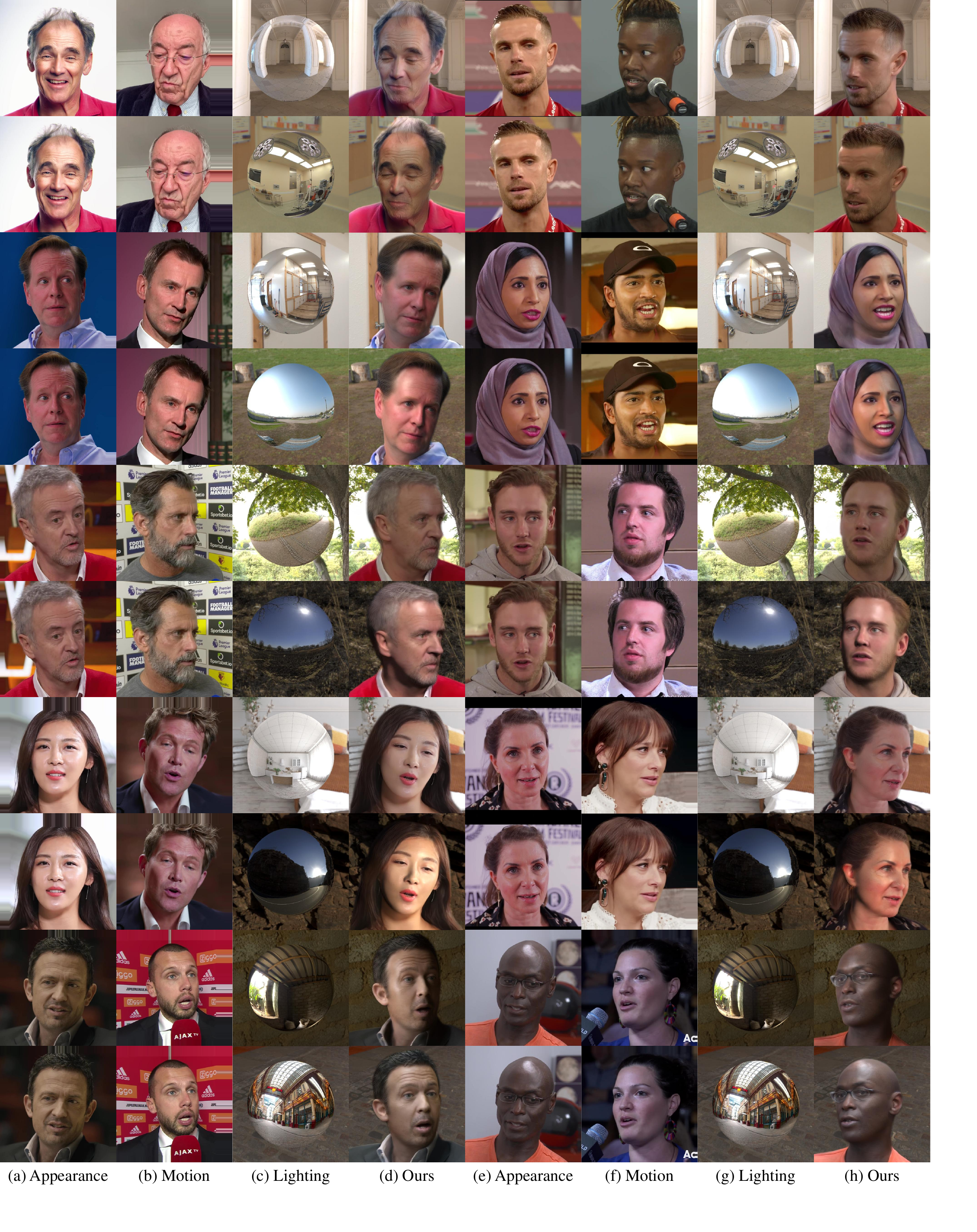}
    \caption{\textbf{Cross-reenactment results on the VFHQ dataset with HDR environment maps as lighting sources.}}
    \label{fig:cross_vfhq_hdri}
\end{figure*}

\begin{figure*}[t]
    \centering
    \includegraphics[width=0.97\linewidth]{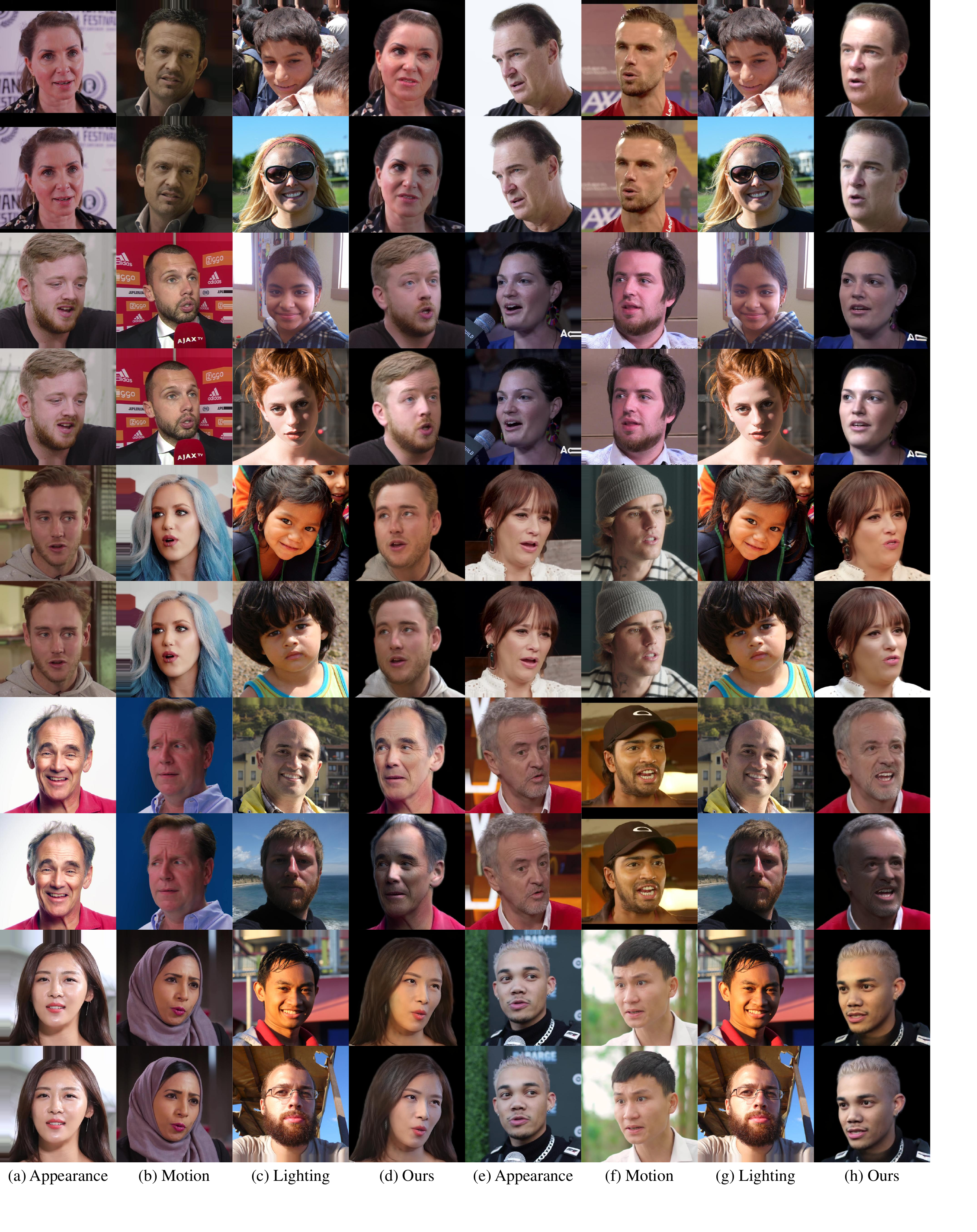}
    \caption{\textbf{Cross-reenactment results on the VFHQ dataset with portrait images as lighting sources.}}
    \label{fig:cross_vfhq_ffhq}
\end{figure*}

We present additional face reenactment results on the VFHQ~\cite{xie2022vfhq} and HDTF~\cite{zhang2021flow} datasets in~\cref{fig:additional_cross_vfhq,fig:cross_hdtf_p1,fig:cross_hdtf_p2,fig:self_hdtf}, demonstrating the effectiveness of~\fullmodel in both motion and lighting control.  
Further, we explore two downstream applications.  
As shown in~\cref{fig:cross_vfhq_hdri}, \fullmodel enables relighting of animated portraits using HDR environment maps, producing a background replacement effect.  
In~\cref{fig:cross_vfhq_ffhq}, \fullmodel can leverage arbitrary portrait images as lighting sources, vividly transferring the illumination effect from one portrait to another.

\section{Dataset Details}
\begin{figure*}[t]
    \centering
    \includegraphics[width=0.97\linewidth]{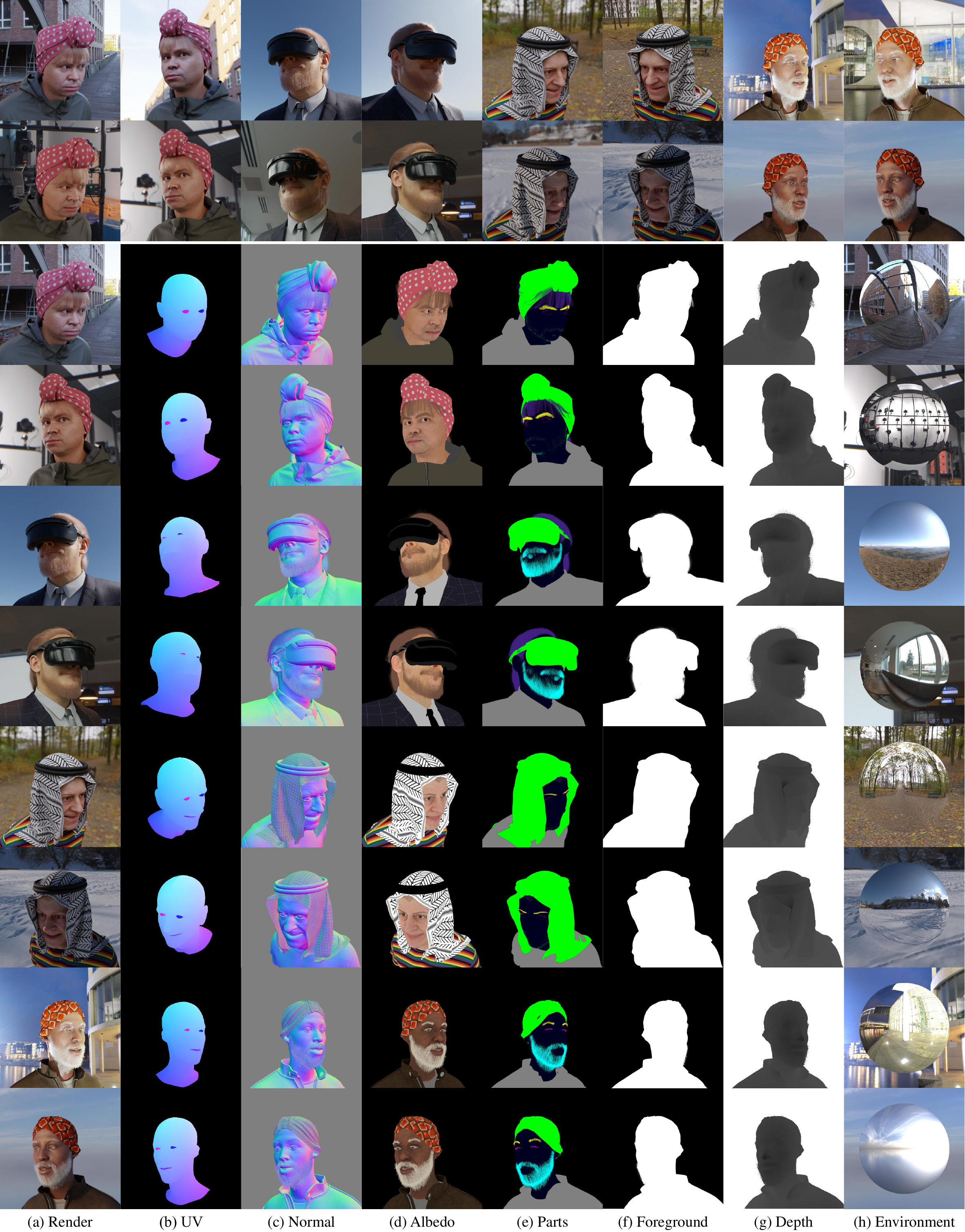}
    \caption{\textbf{Multi-view subset of our synthetic data.} It incorporates 50K subjects. Each is rendered in 2 environments with 10 views.}
    \label{fig:synthetic_multiview}
\end{figure*}

\begin{figure*}[t]
    \centering
    \includegraphics[width=0.97\linewidth]{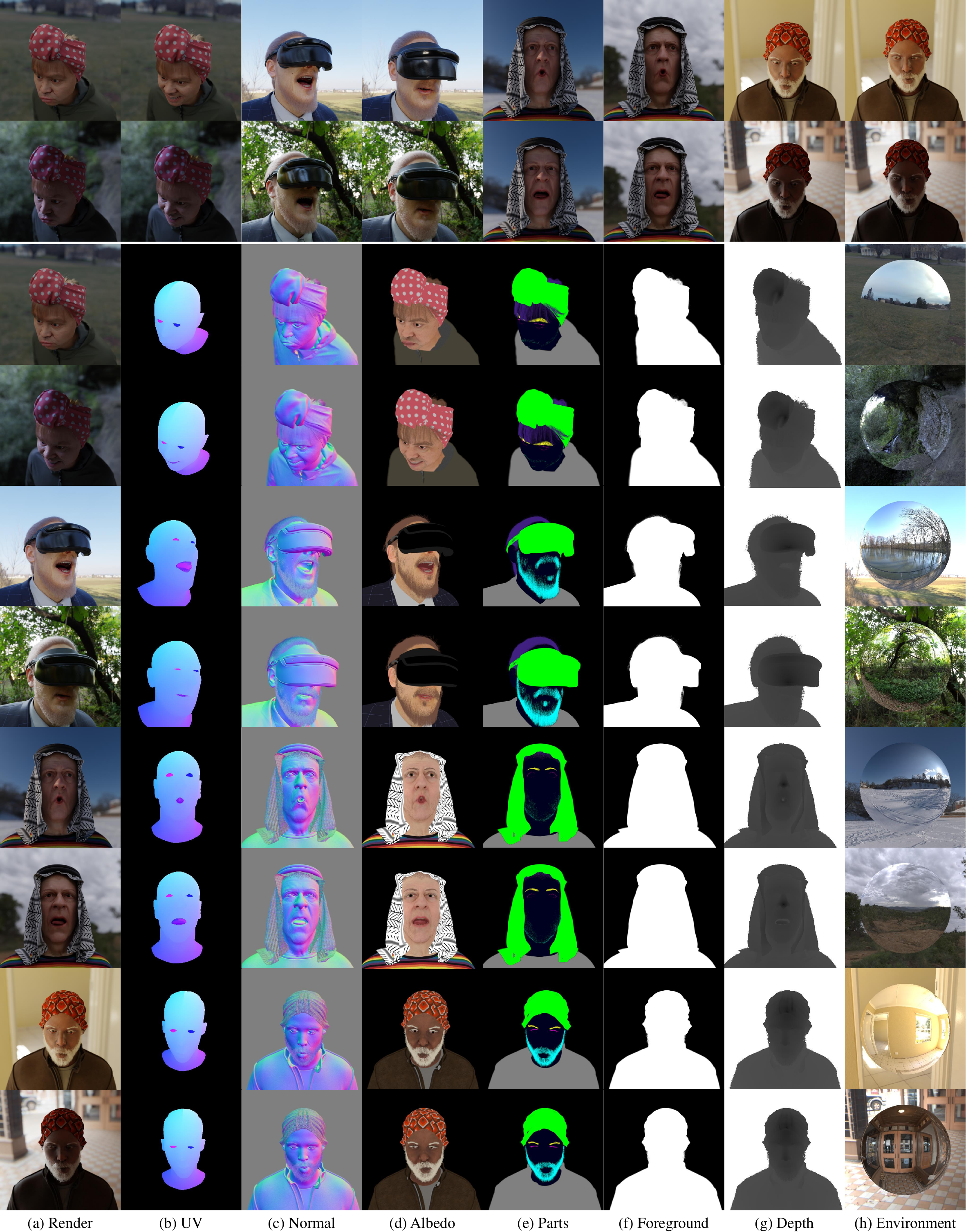}
    \caption{\textbf{Video subset of our synthetic data.} It includes 10K subjects. Each is rendered in 10 environments with 10 poses/expressions.}
    \label{fig:synthetic_video}
\end{figure*}

\begin{figure*}[t]
    \centering
    \includegraphics[width=0.97\linewidth]{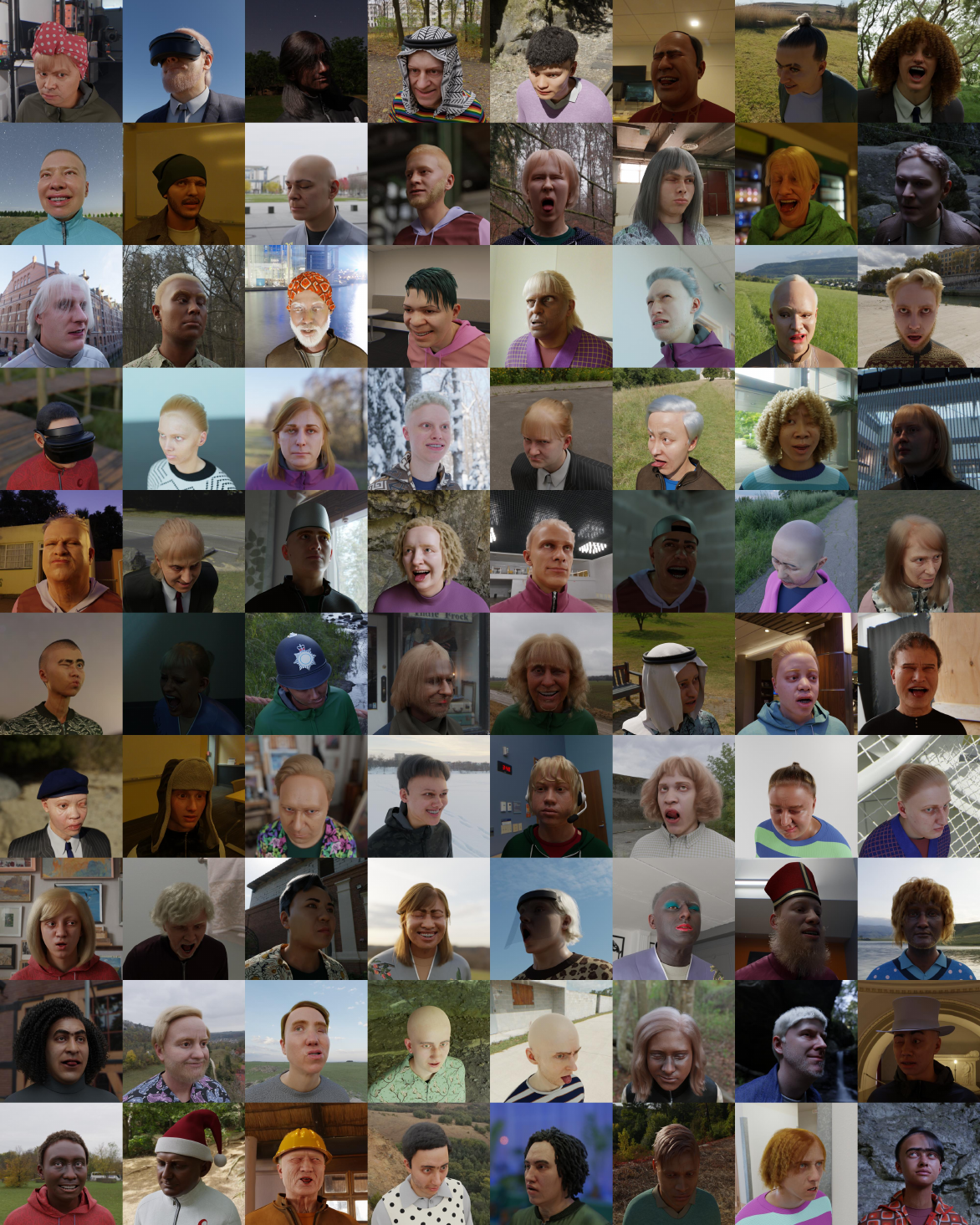}
    \caption{\textbf{More subjects in our synthetic data.} Subjects are with randomized poses, expressions, hairstyles, skin types, accessories, etc.}
    \label{fig:synthetic_dataset}
\end{figure*}

Our synthetic dataset is designed to advance research in general portrait editing, offering two primary subsets: a multi-view subset and a video-like subset.
\begin{itemize}
    \item The multi-view subset comprises 50K subjects, each captured in two distinct environments and viewed from 10 camera angles. This subset provides extensive data for analyzing objects from diverse perspectives and ensuring multi-view consistency. Samples are shown in~\cref{fig:synthetic_multiview}.
    \item The video-like subset includes 10K subjects, each rendered across 10 environments with varied poses and expressions, making it well-suited for studying motion and temporal changes. Samples are shown in~\cref{fig:synthetic_video}.
\end{itemize}
With diverse samples demonstrated in~\cref{fig:synthetic_dataset}, our synthetic dataset consists of 50K subjects and 2M images. It is enriched with ground truth albedo, normal, depth, UV maps, segmentation masks, and HDR environment maps. This dataset addresses critical limitations compared to the existing synthetic datasets. For example, as shown in~\cref{fig:lumos_dataset}, the Lumos dataset~\cite{yeh2022learning} captures each subject only from one view, limiting it to tasks like portrait relighting. In contrast, our dataset incorporates multiple viewpoints and subject movements, better simulating real-world spatiotemporal variations. These improvements make our dataset more versatile and effective for downstream applications requiring spatial and/or temporal coherence.

\section{Evaluation Details}
In~\cref{tab:main}, we exclude both generated and ground truth backgrounds from metric calculations, focusing solely on the quality of the portrait regions. Since the relighting method proposed by Cai et al.~\cite{cai2024real} requires additional cropping for input images, we recompose the outputs with original inputs in~\cref{fig:qualitative_cross} for visual consistency. During quantitative comparisons in~\cref{tab:main}, we only consider the valid areas after cropping. In~\cref{fig:qualitative_relighting}, the backgrounds of 2\supnd to 3\suprd columns are synthesized by inpainting the lighting source portrait using~\cite{suvorov2022resolution}, while those in the 4\supth to 5\supth columns are rendered from the corresponding HDR environment maps.

\end{document}